\documentclass[runningheads]{llncs}

\usepackage{graphicx}
\usepackage{comment}
\usepackage{amsmath,amssymb}
\usepackage{color}
\usepackage{url}
\usepackage{hyperref}
\usepackage{multirow}
\usepackage[Export]{adjustbox}
\usepackage{tabularx}
\usepackage{float}
\usepackage{subfigure}
\usepackage{arydshln}
\usepackage{booktabs}

\newif\ifreview
\reviewfalse

\ifreview
	\usepackage{lineno}

	\linenumbers
\fi

\begin{document}


\def\SubNumber{67}

\def\GCPRTrack{Fast Review Track}

\title{Exploring Seasonal Variability in the Context of Neural Radiance Fields for 3D Reconstruction on Satellite Imagery}

\ifreview
\else

	\author{Liv Kåreborn\inst{1,2}\orcidID{0009-0004-8576-337X} \and
	Erica Ingerstad\inst{1}\and \\
	Amanda Berg\inst{1,3}\orcidID{0000-0002-6591-9400} \and
    Justus Karlsson\inst{1,3}\orcidID{0009-0000-4169-9768} \and
    Leif Haglund\inst{1,3}\orcidID{0009-0006-0801-2024}}
	
	\authorrunning{L. Kåreborn et al.}
	
	\institute{Maxar International Sweden AB, Ebbegatan 13, 582 13 Linköping, Sweden
	\email{\{amanda.berg,leif.haglund\}@maxar.com}\\ 
    \and AI Sweden, Folkungagatan 44, 118 26 Stockholm, Sweden\\
	\email{liv.kareborn@ai.se}\\ 
    \and Computer Vision Laboratory, Department of Electrical Engineering, Linköping University, 581 83 Linköping, Sweden\\
	\email{\{amanda.berg,justus.karlsson,leif.haglund\}@liu.se}}
\fi

\maketitle              

%
%
%


\begin{abstract}
     In this work, the seasonal predictive capabilities of Neural Radiance Fields (NeRF) applied to satellite images are investigated. Focusing on the utilization of satellite data, the study explores how Sat-NeRF, a novel approach in computer vision, performs in predicting seasonal variations across different months. Through comprehensive analysis and visualization, the study examines the model's ability to capture and predict seasonal changes, highlighting specific challenges and strengths. Results showcase the impact of the sun direction on predictions, revealing nuanced details in seasonal transitions, such as snow cover, color accuracy, and texture representation in different landscapes. Given these results, we propose Planet-NeRF, an extension to Sat-NeRF capable of incorporating seasonal variability through a set of month embedding vectors. Comparative evaluations reveal that Planet-NeRF outperforms prior models in the case where seasonal changes are present. The extensive evaluation combined with the proposed method offers promising avenues for future research in this domain.

\end{abstract}    
\section{Introduction}
\label{sec:intro}

Earth observation satellites are essential for monitoring the environment, and the amount of available data increases rapidly. Beneficial applications are found in e.g. meteorology \cite{meteorology}, agriculture \cite{agriculture2}, forestry \cite{forestry_maxar}, biodiversity conservation \cite{bioconservation}, and regional planning \cite{Infrastructure}. As technology advances and the number of orbiting satellites increases, their significance grows \cite{EO_overview}. However, the world we observe through earth observation satellites is far from static, and using these images for 3D reconstruction faces challenges. Due to  variations, including seasonal changes, varying incoming light, and the presence of transient objects, multi-date imagery is frequently disregarded in novel view synthesis. At the same time, these extensive collections of images contain valuable information about how specific regions change over time.

Recent advances in neural 3D reconstruction are neural radiance fields (NeRF) \cite{nerf}, a technique that has demonstrated impressive capabilities of creating accurate representations of 3D objects or scenes by representing them as a 5D-vector valued function or field. This function implicitly handles tasks such as synthesizing new views and reconstructing 3D information, matching geometry and colors with the camera projections from various angles. Since the initial research in this area \cite{nerf}, numerous researchers have explored different aspects \cite{s-nerf,sat-nerf,eo-nerf,nerf-w,sat-mesh}. Some have investigated how this approach copes with transient objects, such as cars \cite{nerf-w}, while others have explored its application to multi-date satellite images \cite{s-nerf,sat-nerf,eo-nerf,sat-mesh}.

Although previous works adapt the use of NeRF to satellite imagery by handling the occurrence of transient objects, an intriguing yet underexplored terrain lies in understanding its capabilities within dynamic environments subject to seasonal changes. Handling images from various seasons can pose challenges in creating an accurate 3D representation of the scene since the seasons contribute to changes in geometry, color, and illumination. Examples of variations include, among other things, varying amounts of snow, and the color and density appearance of vegetation in different seasons. Nevertheless, in the specific case of 3D reconstruction of real-world scenes captured by satellites, the handling of seasonal variability is very important. First of all, for accurate 3D reconstruction in the case of a sparse set of images where season changes between images are present. Second, for the purpose of novel view synthesis with different seasons.

The contribution of this paper is twofold. First, an in-depth evaluation of the seasonal predictive capability of a publicly available method for satellite NeRF 3D reconstruction, Sat-NeRF \cite{sat-nerf}. Second, an extension to Sat-NeRF, called \textit{Planet-NeRF}. A modular, simple, yet effective way of encoding seasons.

\section{Related work}
\label{sec:related_work}

Neural Radiance Fields (NeRF), first presented by  Mildenhall et al. \cite{nerf} in 2020, are capable of generating novel views of complex 3D scenes from 2D input images. Since the method was introduced it has been further developed by multiple researchers to handle e.g. sparse inputs \cite{regnerf,pixelnerf}, slow training time \cite{eff-nerf,eff-nerf2,eff-nerf3}, large scale views \cite{meganerf}, satellite imagery \cite{sat-nerf,eo-nerf}, or inconsistent lighting \cite{s-nerf,nerf-w}. 

\subsection{NeRF applications to satellite imagery}
One major limitation of the original NeRF pipeline \cite{nerf} is that it only uses input images taken around the same time and with similar lighting to get an accurate representation of the scene. This implies challenges when presented with images that contain e.g. dynamic objects, seasonal variations, or shadows, all present in satellite images. These problems have, in recent years, been addressed in multiple works \cite{s-nerf,sat-nerf,eo-nerf,sat-mesh,behari2023sundial,Lv2023}. The first attempt of applying NeRF to satellite images was S-NeRF \cite{s-nerf}, published by Dereksen and Izzo in 2021. They proposed to estimate a sun-dependent shading scalar across images. Marí et al. \cite{sat-nerf} then proposed Sat-NeRF \cite{sat-nerf}, extending S-NeRF by modeling transient features in addition to shadows as well as  representing the input cameras using the RPC model instead of a pinhole camera. Shadows were also addressed by the same authors in EO-NeRF \cite{eo-nerf}, where a geometric NeRF-based shadow rendering approach was proposed.

Common to all of the works mentioned above, is that they all evaluate on datasets free from seasonal changes. In this work, the seasonal predictive capability of Sat-NeRF is evaluated and the connection to the sun direction explained. We then propose Planet-NeRF, an extension to Sat-NeRF, capable of incorporating seasonal variability through month embedding vectors. EO-NeRF \cite{eo-nerf}, has no publicly available source code, neither have they published any results on any data set containing seasonal variation.

\subsection{Seasonal variability}
In this work, seasonal variability refers to the distinct environmental changes throughout the four seasons, such as snowy winters and green summers. To the best of our knowledge, the only previously published methods that address the problem of seasonal variability in the context of satellite imagery is Season-NeRF  \cite{season-nerf} and Sat-mesh \cite{sat-mesh}. Season-NeRF employs a NeRF with time-of-year-based adjustments for seasonally adjusted albedo colors. 
A height map is introduced for early training guidance. Their approach differs from ours as we compute an albedo color based on only the spatial coordinates and then also separately compute a seasonal color.

Instead of predicting a density sigma, Sat-Mesh \cite{sat-mesh} predicts a signed distance function and the color prediction is extended in order to learn image-specific appearances, for example seasons. A latent embedding vector is incorporated in the training process to encode every image appearance. The method allows the model to learn and render the different seasons and enables to texture the mesh with the corresponding seasonal appearances of each image. Unlike our latent vectors, theirs is specific to each image, while we generate the latent appearance for each month. Sat-Mesh does not provide source code or results in SSIM, PSNR, or altitude MAE for the Omaha areas and, therefore, can not be compared to.
\section{Methodology}
\label{sec:methodology}

Seasons exist because of Earth's tilted axis and they are correlated to the solar paths. Section \ref{sec:equinoxes} introduces the reader to the concept of equinoxes, which is key to understanding why Sat-NeRF fails in accurately representing different seasons. Following this introduction, the proposed extension Planet-NeRF is presented in Section \ref{sec:MEV}.

\subsection{Scientific correlation between solar paths and seasons}\label{sec:equinoxes}

The Earth's axis is inclined at approximately 23.5 degrees relative to its orbital plane around the Sun, causing the cyclical transition of seasons throughout the year. 
Equinoxes, such as the vernal equinox around March 21st and the autumnal equinox around September 23rd, mark times when the Earth's axis is neither tilted towards nor away from the Sun. During these equinoxes, day and night exhibit approximate equality in length globally and the Sun's path observed from Earth remains consistent, making it identical for both spring and fall \cite{cite-sun-path}, while the seasonality appears differently.

\subsection{Planet-NeRF}\label{sec:MEV}
\begin{figure}[htbp] 
\center{\includegraphics[width=\columnwidth]{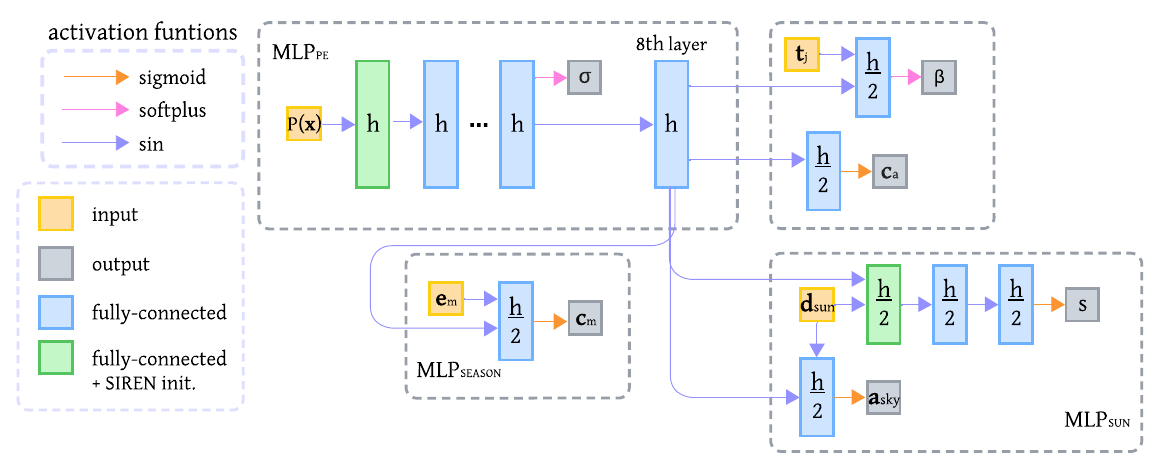}}
\caption{\normalsize{
The Planet-NeRF network is formed by integrating the Sat-NeRF network architecture with an additional layer ($\text{MLP}_{\text{season}}$) representing the latent month embedding vectors $\textbf{e}_m\in\mathbb{R}^K$, where $m \in [1,12]$, used to predict a seasonal albedo color $\textbf{c}_m\in\mathbb{R}^{N_s}$. In contrast to Sat-NeRF, Planet-NeRF also employs positional encoding of the input. $\textbf{t}_j\in\mathbb{R}^4$ is an embedding vector representing transient objects for image $j$, $\sigma$ is the volume density, $\textbf{d}_\text{sun}\in\mathbb{R}^{N_s}$ the sun direction, $s$ the shading scalar, $\beta$ the uncertainty coefficient, $\textbf{c}_\text{a}\in\mathbb{R}^{N_s}$ the albedo prediction, and $\textbf{a}_\text{sky}\in\mathbb{R}^{N_s}$ the ambient sky color prediction. $\textbf{h}\in\mathbb{R}^{512}$ and $x\in\mathbb{R}^{N_s}$. $N_s$ denotes the number of samples along each ray. $N_s=64$ and the dimensions of $\textbf{h}$ and $\textbf{t}_j$ are the default values of Sat-NeRF.
}}
\label{img:sat_and_eo_with_month}
\end{figure}

The proposed method \textit{Planet-NeRF} extends the Sat-NeRF \cite{sat-nerf} architecture by combining two modifications. 1) by the incorporation of a set of month embedding vectors and, 2) by the introduction of a positional encoding of the input. An overview of the architecture can be seen in Figure \ref{img:sat_and_eo_with_month}.

\subsubsection{Month embedding vectors}
In order to teach the model about key characteristics for different months and, therefore, different seasons, a set of month embedding vectors $\textbf{e}_m\in\mathbb{R}^K$, where $m \in [1,12]$, was incorporated in the Sat-NeRF network, see Figure \ref{img:sat_and_eo_with_month}. 
The 12 month embedding vectors are $K$-dimensional, hence, each month is represented by a $K$-dimensional vector that is trained to represent the appearance of the specific month. Furthermore, each month embedding vector is used to predict an month albedo color, $\textbf{c}_m$, which is incorporated into the final color prediction for each image (see Equation \ref{planet_nerf_trans_eq}). $K$ is set to 4, a choice based on initial evaluations conducted during testing. Optimization of the value of $K$ is left for future work.   

\subsubsection{Positional encoding}
In the original NeRF paper \cite{nerf} the authors used positional encoding to enable the network to learn high frequencies of the target image. Positional encoding is an expansion of the 3-dimensional input coordinates into a high-dimensional high-frequency space.

In a follow up paper \cite{tancik2020fourfeat}, the authors showed that an expansion of the input coordinates is indeed \emph{necessary} for the MLP to learn high frequencies in the target image. In its default configuration, Sat-NeRF does not use positional encoding, and instead feeds the raw 3D coordinates into the MLP. A positional encoding of 10 frequencies was added to the Planet-NeRF model for evaluation. Season-NeRF also uses positional encoding, though the number of frequencies is not specified. 

The following equation was used for positional encoding, where $\textbf{x}$ is the 3-dimensional input coordinates and $N_f$ the number of frequencies: 
\begin{equation}
    \begin{split}
    P(\textbf{x}) = [\dots , \sin (2^k \pi \textbf{x}) , \cos (2^k \pi \textbf{x}) , \dots], \\ 
    \text{ for } k = 0, \dots, N_f - 1.
    \end{split}
    \label{eq:postitional_encoding}
\end{equation}



 \subsubsection{Color calculation}\label{seq:rendering_function}
The color $\textbf{c}_i$ is calculated as follows for Planet-NeRF:
\begin{equation}
    \textbf{c}_i = \textbf{c}_m \textbf{c}_\text{a} \cdot (\text{s} + (1 - \text{s}) \textbf{a}_\text{sky})
    \label{planet_nerf_trans_eq}
\end{equation}
where $\textbf{c}_m$ denotes the seasonal albedo color,  $\textbf{c}_\text{a}$ the albedo prediction, $s$ the shading scalar, and $\textbf{a}_{sky}$ the ambient sky color prediction.

\subsubsection{Training methodology}
During the training process, the embedding vectors $\textbf{e}_m$ are embedded into the final layer of the MLP but are not integrated until the third epoch. The decision to wait for three epochs is based on extensive testing, which showed that this approach yielded superior results in terms of PSNR, SSIM, and altitude MAE.
Each image in the data sets is labelled with the date it was captured. During training, an image captured in month $m$ updates the corresponding month's embedding vector, $\textbf{e}_m$. During inference, images are textured using the embedding vector $\textbf{e}_m$ from the month they were captured.

\section{Experiments}
\label{sec:experiments}
In this section, details of the conducted experiments are presented. An ablation study of the proposed method is performed, as well as a comparison to Season-NeRF \cite{season-nerf}. PSNR, SSIM, and altitude MAE for Season-NeRF were sourced from their paper, specifically the results from their full model. Sat-NeRF results were obtained by downloading and running the default configuration from \cite{satnerfgithub}.

\subsection{Data sets}
The evaluation in this study utilizes data sets from the \textit{2019 IEEE GRSS Data Fusion Contest} \cite{data}, encompassing eight distinct areas located in Omaha, Nebraska, USA, and two areas from Jacksonville, Florida, USA. All images employed are sourced from Maxar WorldView-03, possessing an image size of 2048x2048 pixels and covering an approximate area of 25000 $\text{m}^\text{2}$. Details are presented in Table \ref{tab:datasets}. In order to optimize training time, all images are down-sampled by a factor of four, resulting in a resolution of 512x512 pixels for training purposes. The Omaha data sets include seasonal variations, while Jacksonville do not.

\begin{table}[]
\begin{center}
\caption{Number of train $N_{tr}$ and test images $N_{te}$, and height range $r_h$ for all of the data sets utilized in the experiments.}
\begin{tabular}{ |l||c|c|c|c|c|c|c|c|c|c| }
 \hline
 & \multicolumn{2}{c|}{\textbf{JAX}} & \multicolumn{8}{c|}{\textbf{OMA}} \\
 \hline
 ID & 004 & 068 & 042 & 084 & 132 & 163 & 203 & 212 & 281 & 374 \\
 \hline
 $N_{tr}$ & 9 & 17 & 35 & 22 & 37 & 36 & 39 & 34 & 37 & 26  \\
 \hline
 $N_{te}$ & 2 & 2 & 4 & 4 & 4 & 4 & 4 & 4 & 4 & 4  \\
 \hline
 $r_h$ [m] & 25.0 & 57.0 & 30.0 & 53.0 & 31.9 & 32.0 & 44.0 & 34.6 & 60.0 & 59.4 \\
 \hline
\end{tabular}
\label{tab:datasets}
\end{center}
\end{table}

The ground truth for these data sets consists of both RGB images, akin to those used for training, and lidar data. The lidar data exhibits a resolution of 0.3 m/pixel at nadir. It is important to note that the lidar data has been collected at a specific point in time, whereas the satellite images has been captured during an extended period. Hence, in areas that have dynamic objects and/or areas that exhibit large differences due to seasonal variations, e.g. areas with vegetation and/or deep snow, the ground truth depth do not necessarily reflect the true depth. The lidar data is, however, the best approximation of the true depth available.

For the Jacksonville data sets, the training and test image split aligns with the methodology adopted by Sat-NeRF. In the case of Omaha, the number of training images per month is unevenly distributed and varies between 0 and 10 depending on availability, for more detailed information, see \cite{data}. The four test images are selected to ensure representation of each distinct season. Test images from Jan, Mar, Sep, and Oct are used for all OMA areas except 084 that have April instead of March and 212 that have images from Jan, Mar, Apr, and Sep. The reason for this being the unavailability of images for all months for some data sets. The number of days between the test and train images differs for all of the data sets and test images. They do, however, all have a corresponding image in the training data for that month.

\subsection{Training}
\label{sec:training}
Consistent parameters across all models include network parameters, activation functions, number of layer samples, and batch size 
mirroring those employed by the default configuration of Sat-NeRF. The training time was determined through trial experiments that monitored convergence. The training time for all models was, thereafter, standardized to 20 epochs.
\section{Results and discussion}
\label{sec:results_and_discussion}

\subsection{Sat-NeRF and seasonal variability} \label{sec:result-RQ1-satnerf}
In this section, the results of Sat-NeRF on different areas in Omaha, that experience seasonal variability, are compared with the results from Jacksonville, a location characterized by minimal seasonal variations.

\subsubsection{Qualitative and quantitative results}
In terms of PSNR and SSIM scores, Sat-NeRF achieves better results for Jacksonville than Omaha areas. Mean PSNR for the two Jacksonville areas is 22.64 while 17.93 for Omaha, and mean SSIM for Jacksonville is 0.82 while 0.52 for Omaha (higher is better for both metrics). Mean altitude MAE is 3.05 for the Omaha areas compared to 2.01 for Jacksonville (lower is better). Notably, the quantitative results for Jacksonville is better for all metrics compared to Omaha even though the Omaha data sets have roughly twice the number of training images (see Table \ref{tab:datasets}).

\begin{figure}[h]
  \setlength\tabcolsep{2pt}
  \adjustboxset{width=\linewidth,valign=c}
  \centering
  \begin{tabularx}{0.7\linewidth}{@{}
      l
      X @{\hspace{1pt}}
      X @{\hspace{1pt}}
      X @{\hspace{1pt}}
      X @{\hspace{1pt}}
      X 
    @{}}
    & \multicolumn{1}{c}{\textbf{SN}}
    & \multicolumn{1}{c}{\textbf{ME}}
    & \multicolumn{1}{c}{\textbf{PE}}
    & \multicolumn{1}{c}{\textbf{PN}}
    \\
    \rotatebox[origin=c]{90}{$\textbf{c}_\text{a}$}
    & \includegraphics{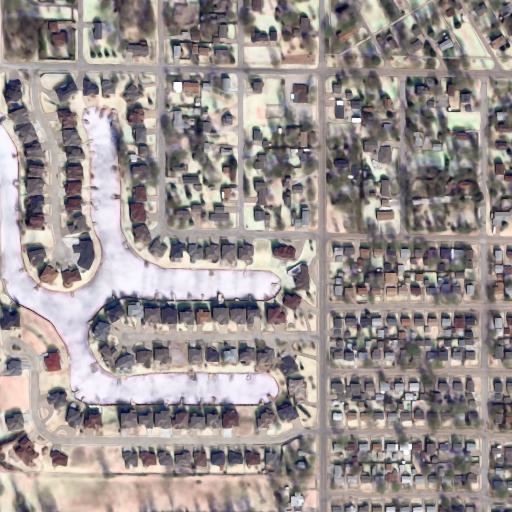}
    & \includegraphics{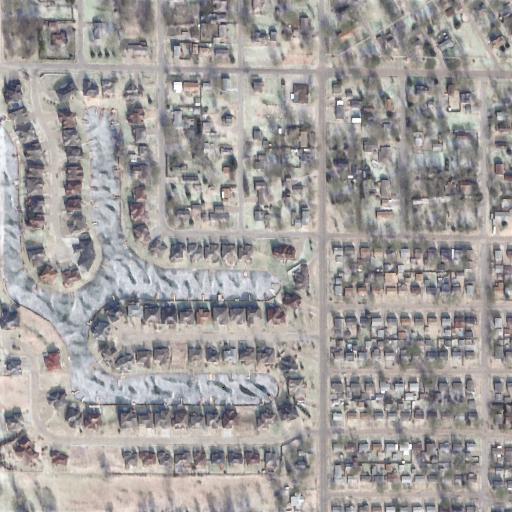}
    & \includegraphics{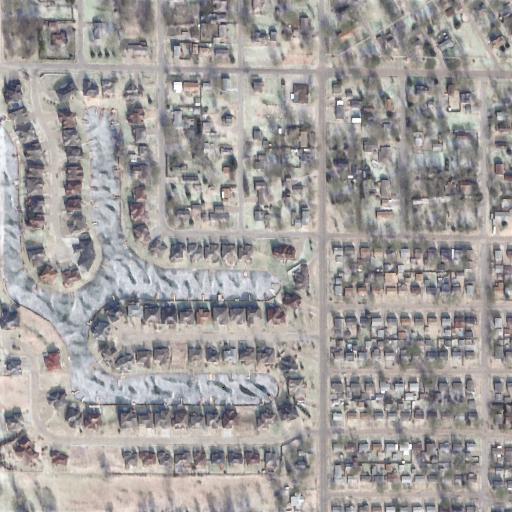}
    & \includegraphics{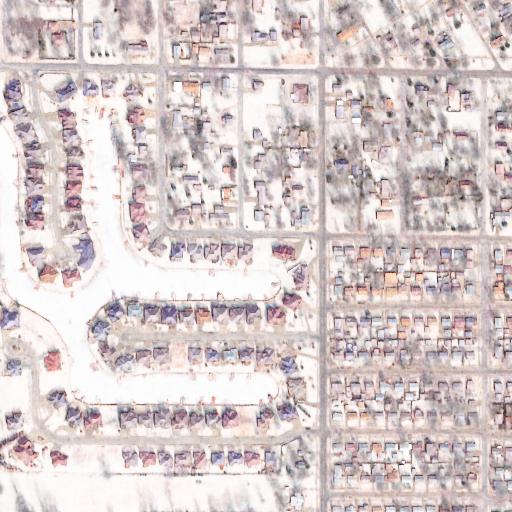}
  \end{tabularx}
\caption{Albedo predictions $\textbf{c}_\text{a}$ for OMA\_132 for models SN: Sat-NeRF, ME: Sat-NeRF + month embedding, PE: Sat-NeRF + positional encoding, and PN: Planet-NeRF.}
\label{fig:albedo}
\end{figure}

Figure \ref{fig:rgb_predictions} presents the rendered RGB images produced by Sat-NeRF in four of the Omaha regions for three different months. These visuals capture the seasonal variations winter, spring, and summer, providing a comprehensive representation of the network's performance across varying seasonal appearances. 
Albedo predictions for area OMA\_132 can be seen in Figure \ref{fig:albedo}. The albedo predictions are visually similar for all Omaha areas, OMA\_132 is included as a representative example.

\begin{figure}[h]
  \setlength\tabcolsep{2pt}
  \adjustboxset{width=\linewidth,valign=c}
  \centering
  \begin{tabularx}{0.6\linewidth}{@{}
      l
      X @{\hspace{1pt}}
      X @{\hspace{1pt}}
      X @{\hspace{1pt}}
      X 
    @{}}
    & \multicolumn{1}{c}{$\textbf{c}_\text{gt}$}
    & \multicolumn{1}{c}{$\textbf{c}$}
    & \multicolumn{1}{c}{$\textbf{c}_\text{a}$}
    \\
    \rotatebox[origin=c]{90}{\textbf{004}}
    & \includegraphics{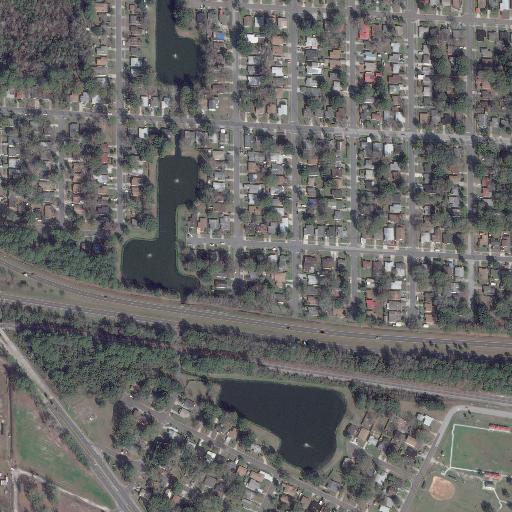} 
    & \includegraphics{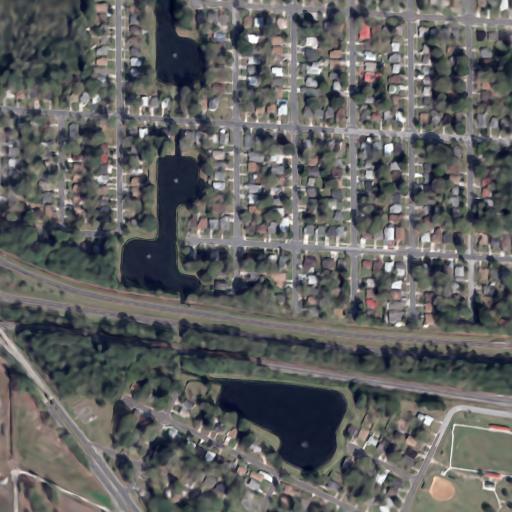}
    & \includegraphics{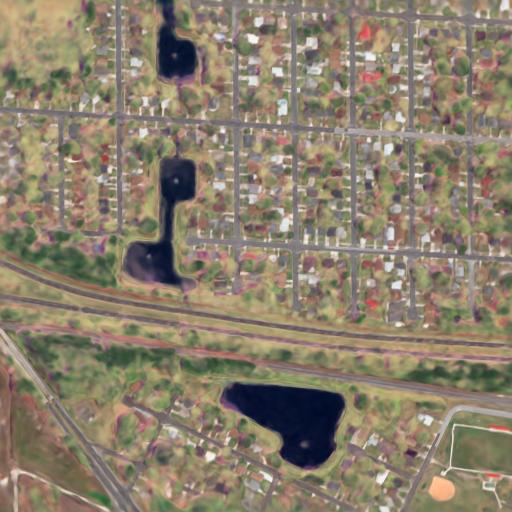}
    \\ \rule{0pt}{36pt}
    \rotatebox[origin=c]{90}{\textbf{068}}
    & \includegraphics{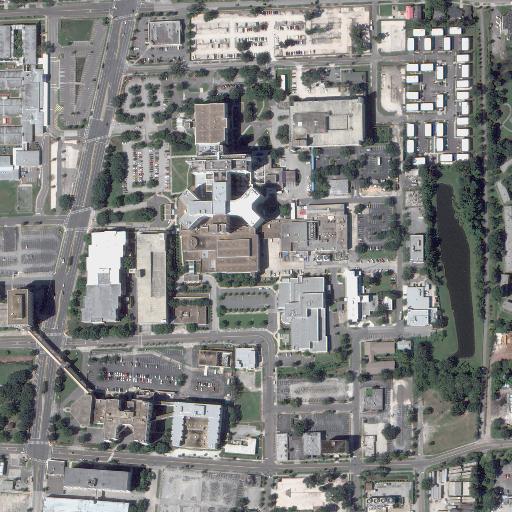} 
    & \includegraphics{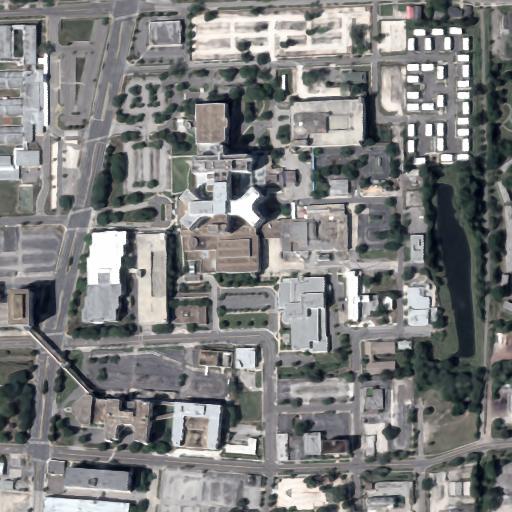}
    & \includegraphics{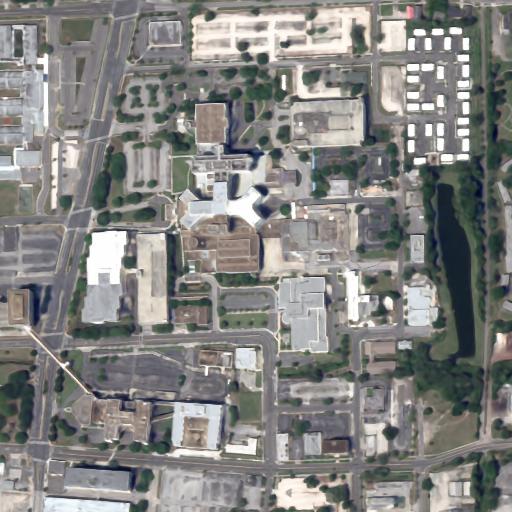}
  \end{tabularx}
\caption{Ground truth $\textbf{c}_\text{gt}$, final RGB prediction $\textbf{c}$, and albedo $\textbf{c}_\text{a}$ predictions by Sat-NeRF for the Jacksonville areas.}
\label{fig:albedo-jax}
\end{figure}



The albedo predictions for Omaha predominantly appear white or pink, deviating significantly from the true scene colors. Conversely, albedo predictions for Jacksonville more closely resemble ground truth colors, as seen in Figure \ref{fig:albedo-jax} for January in JAX\_004 and October in JAX\_068. This likely arises from the fact that the albedo color, which is predicted solely from the position $\textbf{x}$, struggles to understand the significant variations in appearance between seasons for Omaha areas. Conversely, the images from Jacksonville exhibit greater consistency, enabling the model to generate more accurate color predictions based solely on position $\textbf{x}$.

Despite the aforementioned points, it is still clear that Sat-NeRF is successfully capturing a lot of the key characteristics of the different seasons in most of the cases.

\subsubsection{Solar impact of Sat-NeRF predictions}
\begin{figure}[h]
  \setlength\tabcolsep{2pt}
  \adjustboxset{width=\linewidth,valign=c}
  \centering
  \begin{tabularx}{0.8\linewidth}{@{}
      l
      l
      X @{\hspace{1pt}}
      X @{\hspace{1pt}}
      X @{\hspace{4pt}}
      X @{\hspace{1pt}}
      X @{\hspace{1pt}}
      X @{\hspace{1pt}}
      X 
    @{}}
    && \multicolumn{3}{c}{\textbf{January}} & \multicolumn{3}{c}{\textbf{September}}
    \\ \cmidrule(r{8pt}){3-5} \cmidrule{6-8}
    && \multicolumn{1}{c}{\textbf{132}}
    & \multicolumn{1}{c}{\textbf{212}}
    & \multicolumn{1}{c}{\textbf{374}}
    & \multicolumn{1}{c}{\textbf{132}}
    & \multicolumn{1}{c}{\textbf{212}}
    & \multicolumn{1}{c}{\textbf{374}}
    \\ 
    &\rotatebox[origin=c]{90}{$\textbf{c}_\text{gt}$}
    & \includegraphics{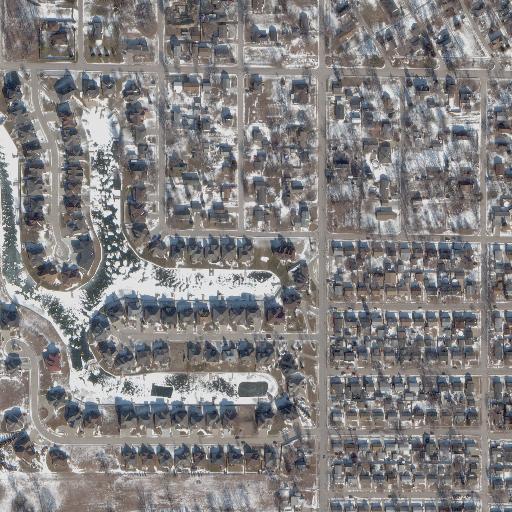}
    & \includegraphics{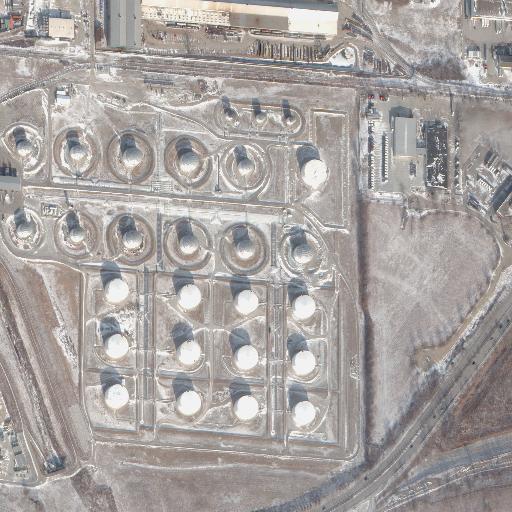}
    & \includegraphics{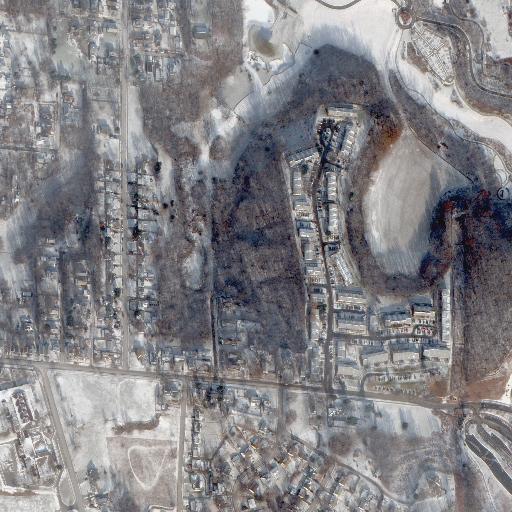}
    & \includegraphics{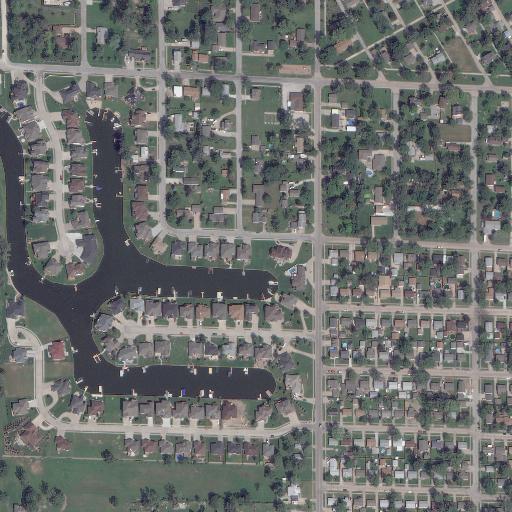}
    & \includegraphics{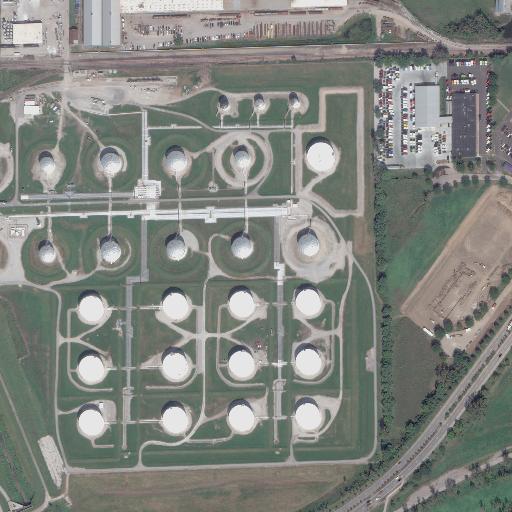}
    & \includegraphics{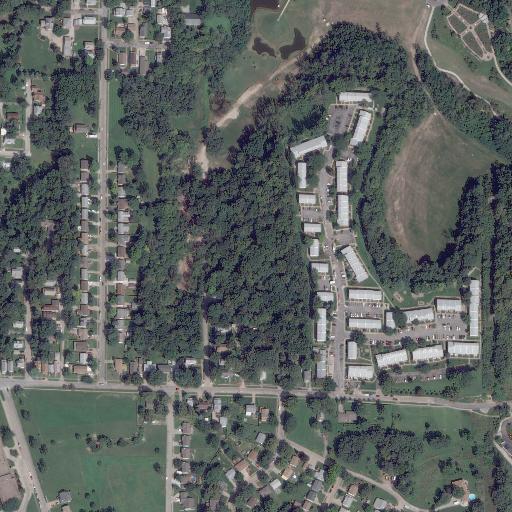}
    \\ \rule{0pt}{24pt}
    \multirow{2}{*}{\rotatebox[origin=c]{90}{\textbf{Sat-NeRF}}}
    & \rotatebox[origin=c]{90}{$s$}
    & \includegraphics{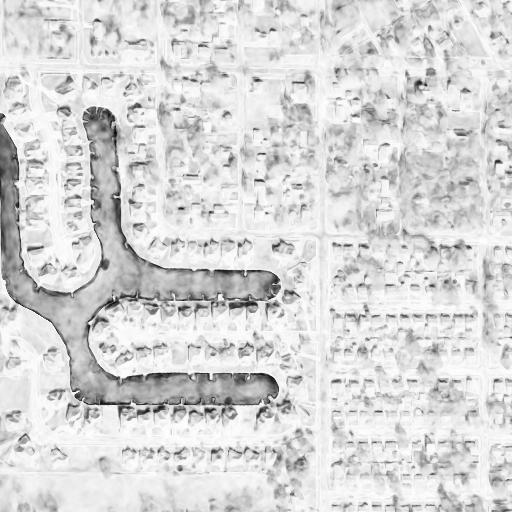}
    & \includegraphics{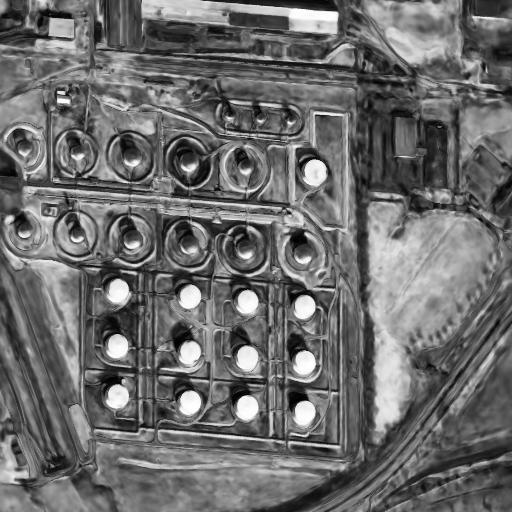}
    & \includegraphics{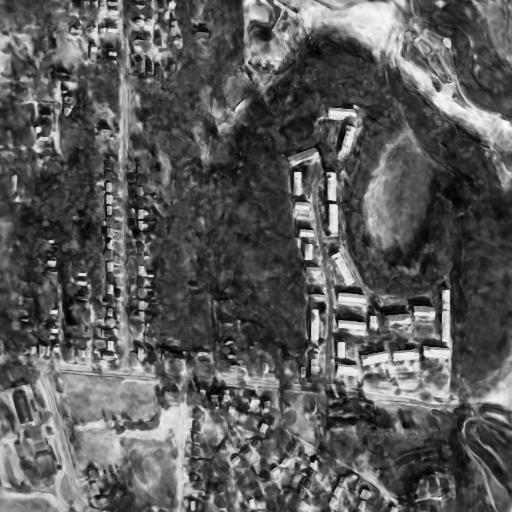}
    & \includegraphics{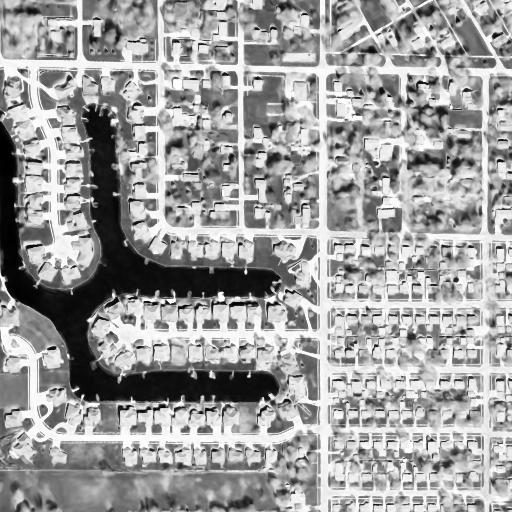}
    & \includegraphics{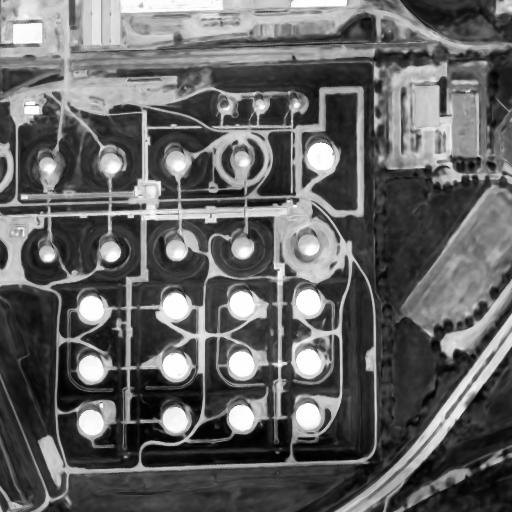}
    & \includegraphics{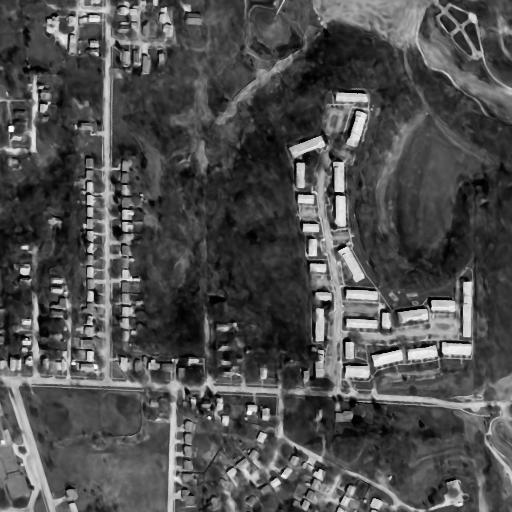}
    \\ \rule{0pt}{24pt}
    & \rotatebox[origin=c]{90}{$\textbf{a}_\text{sky}$}
    & \includegraphics{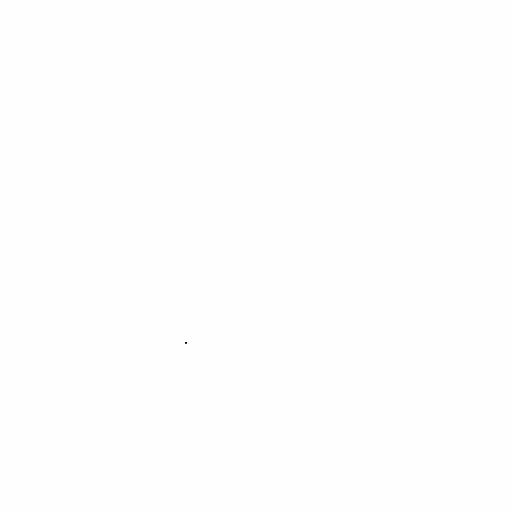}
    & \includegraphics{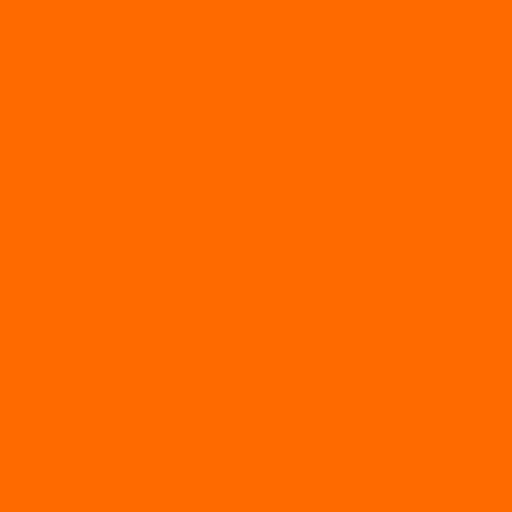}
    & \includegraphics{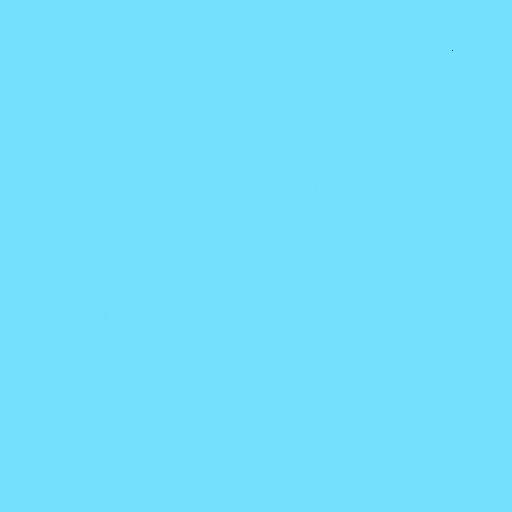}
    & \includegraphics{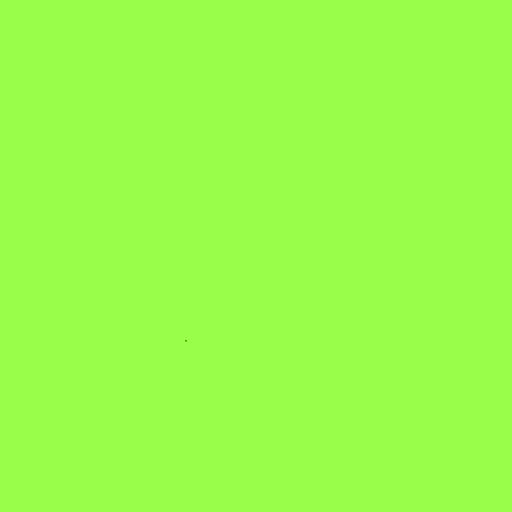}
    & \includegraphics{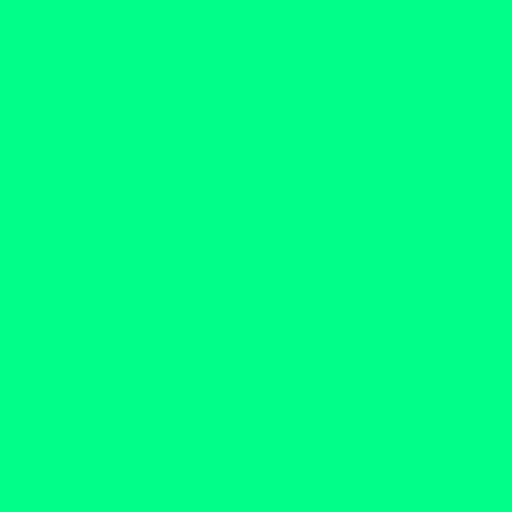}
    & \includegraphics{sec/img/satnerf-sky/OMA_212_036_RGB_epoch20.jpg}
    \\ \rule{0pt}{24pt}
    \multirow{2}{*}{\rotatebox[origin=c]{90}{\textbf{Planet-NeRF}}}
    & \rotatebox[origin=c]{90}{$s$}
    & \includegraphics{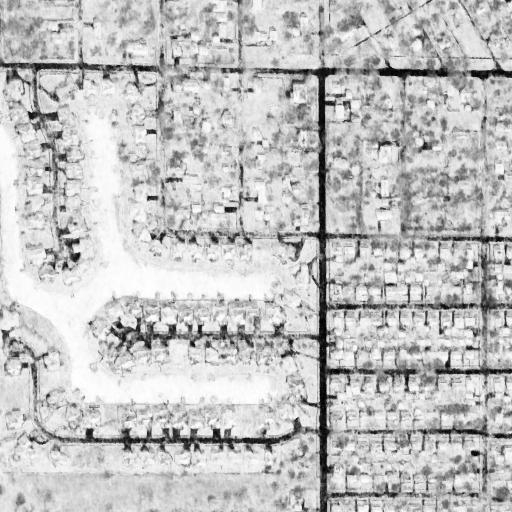}
    & \includegraphics{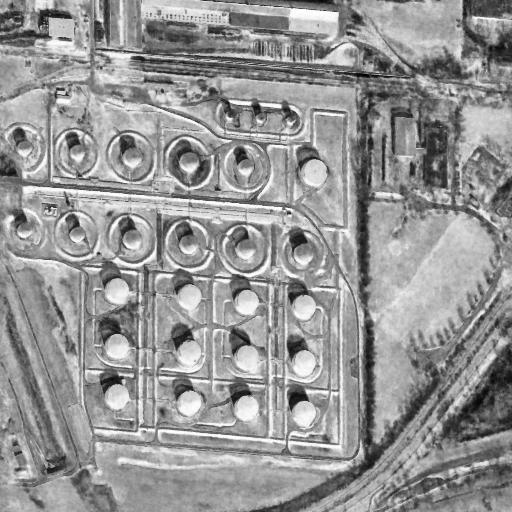}
    & \includegraphics{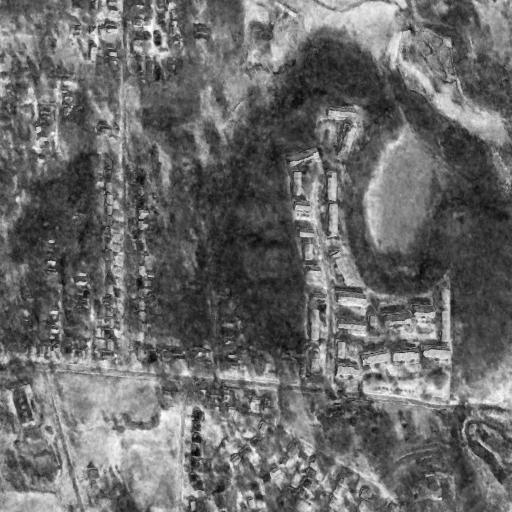}
    & \includegraphics{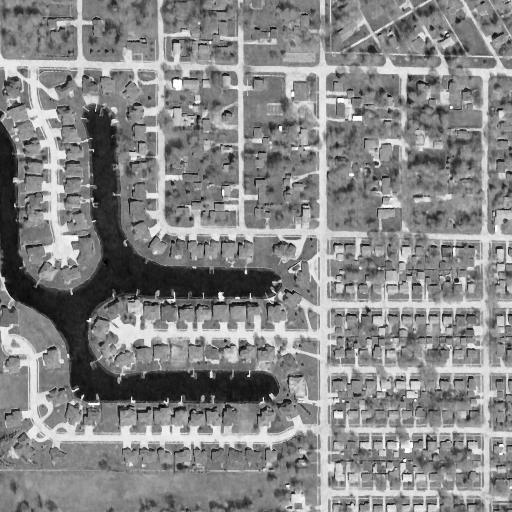}
    & \includegraphics{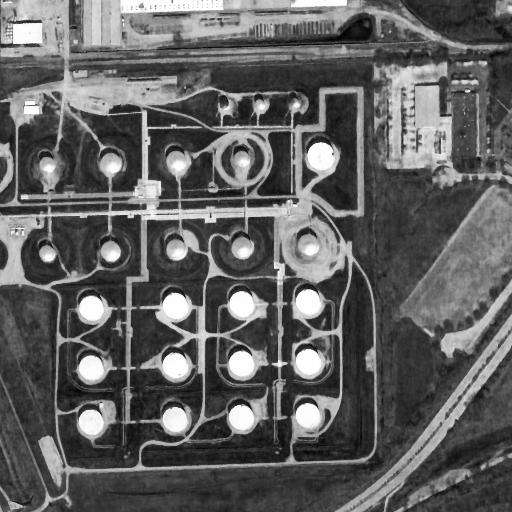}
    & \includegraphics{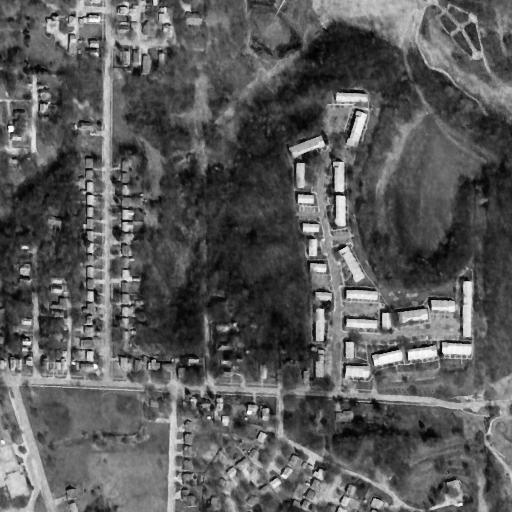}
    \\ \rule{0pt}{24pt}
    & \rotatebox[origin=c]{90}{$\textbf{a}_\text{sky}$}
    & \includegraphics{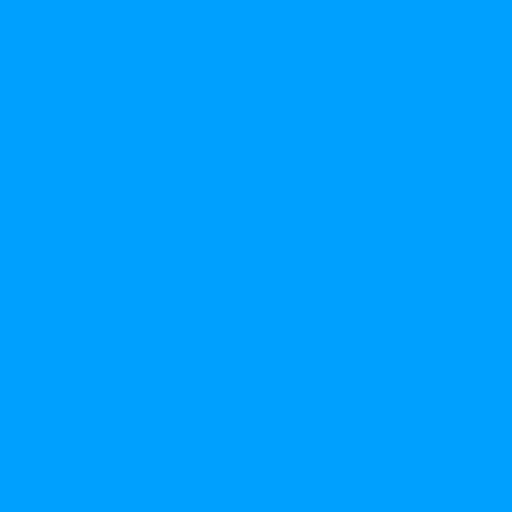}
    & \includegraphics{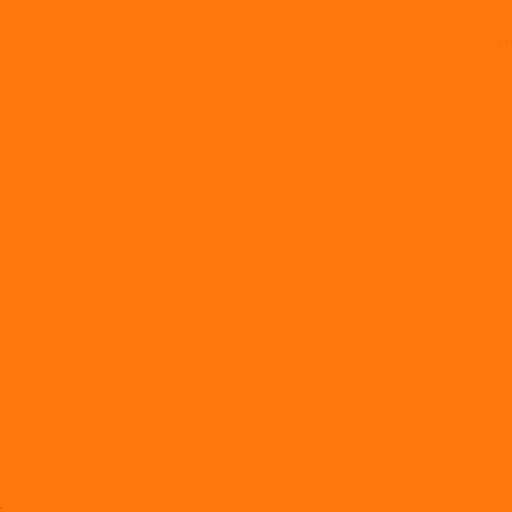}
    & \includegraphics{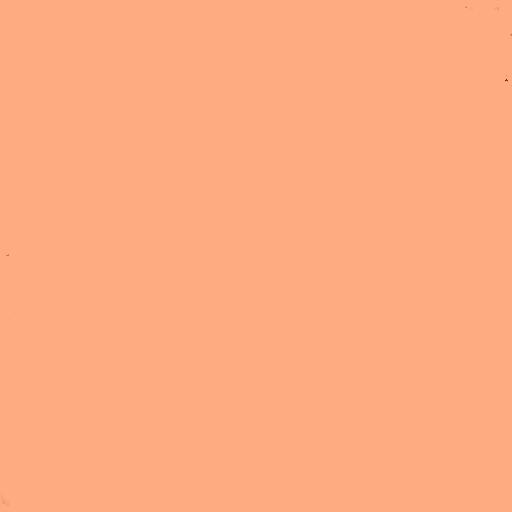}
    & \includegraphics{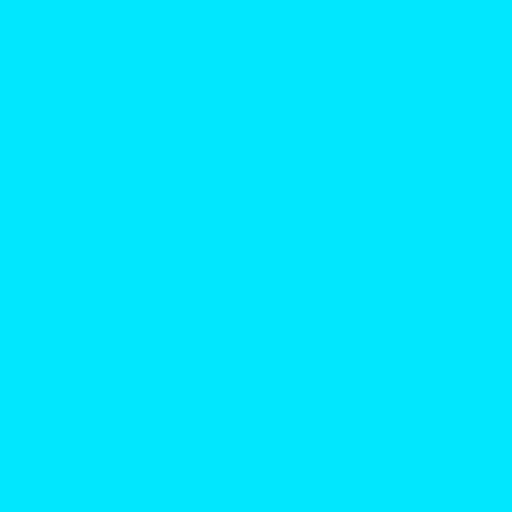}
    & \includegraphics{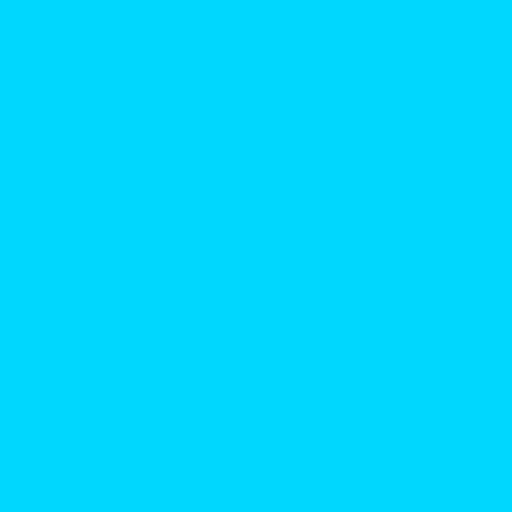}
    & \includegraphics{sec/imgs2/planet-nerf-2/sky/OMA_212_036_RGB_epoch20.jpg}
  \end{tabularx}
\caption{Ground truth $\textbf{c}_{\text{gt}}$, shading scalar $s$, and ambient sky color predictions $\textbf{a}_\text{sky}$ for Sat-NeRF and Planet-NeRF on three different Omaha areas during two distinct seasons.}
\label{fig:sat-nerf-sun-impact-oma}
\end{figure}

In Figure \ref{fig:sat-nerf-sun-impact-oma}, the shading scalar intended to have low values in shadowed areas for applying the ambient sky color also exhibits low values in areas with significant seasonal variations (such as grass and trees). Additionally, the predicted sky color does not seem to resemble typical sky colors; rather, it appears to align more closely with the dominant color in the image. Sat-NeRF's ability to capture essential seasonal characteristics might be attributed to the influence of the sun. While the albedo predictions fail to capture seasonal changes, the final RGB prediction, enhanced by the introduction of sun direction, to some extent succeeds in doing so, see Figure \ref{fig:rgb_predictions} for reference. This suggests that the inclusion of sun direction $\textbf{d}_\text{sun}$ enables Sat-NeRF to grasp seasonal variations, a statement supported by the observation that the shading scalar and sky ambient factors also seem to encapsulate these seasonal changes, as mentioned above.

\begin{figure}[h]
  \setlength\tabcolsep{2pt}
  \adjustboxset{width=\linewidth,valign=c}
  \centering
  \begin{tabularx}{0.6\linewidth}{@{}
      l
      X @{\hspace{1pt}}
      X @{\hspace{1pt}}
      X @{\hspace{1pt}}
      X @{\hspace{1pt}}
      X 
    @{}}
    & \multicolumn{1}{c}{\textbf{004}}
    & \multicolumn{1}{c}{\textbf{004}}
    & \multicolumn{1}{c}{\textbf{068}}
    & \multicolumn{1}{c}{\textbf{068}}
    \\
    \rotatebox[origin=c]{90}{$\textbf{c}_\text{gt}$}
    & \includegraphics{sec/img/JAX/gt_rgb/JAX_004_009_RGB_epoch20.jpg}
    & \includegraphics{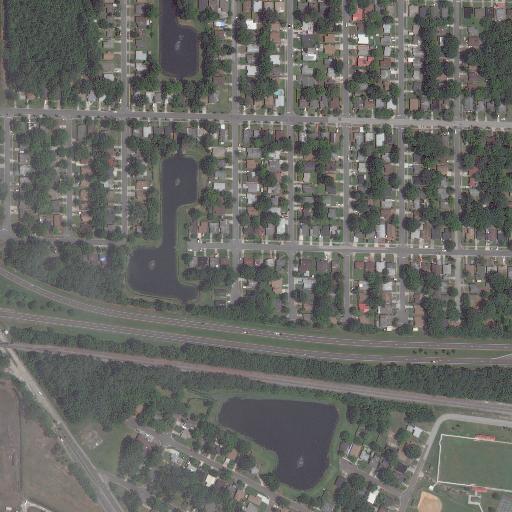}
    & \includegraphics{sec/img/JAX/gt_rgb/JAX_068_002_RGB_epoch20.jpg}
    & \includegraphics{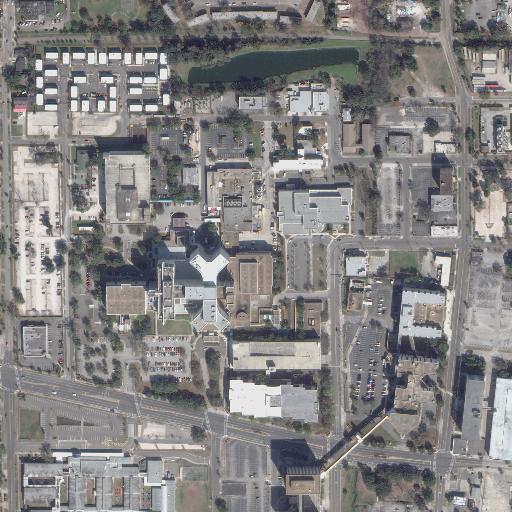}
    \\ \rule{0pt}{27pt}
    \rotatebox[origin=c]{90}{$s$}
    & \includegraphics{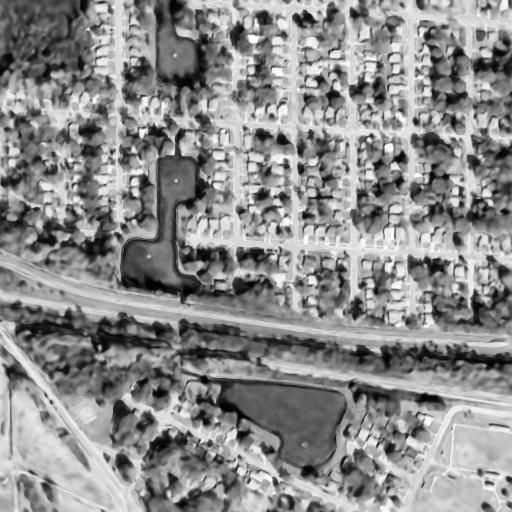}
    & \includegraphics{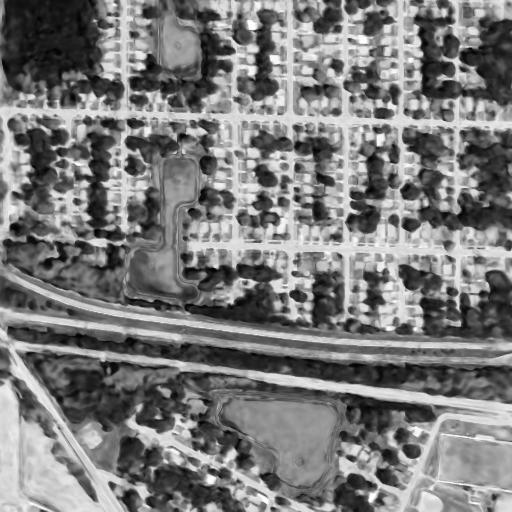}
    & \includegraphics{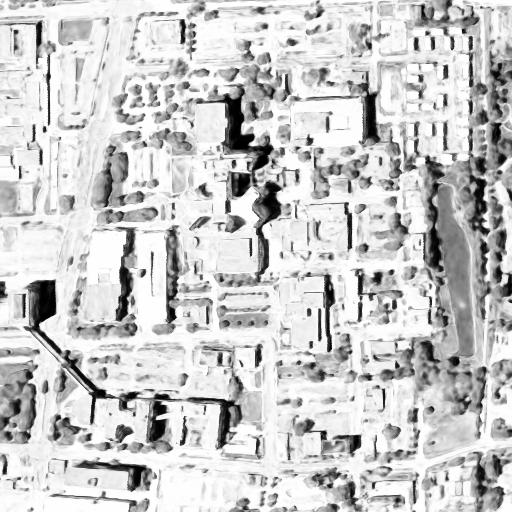}
    & \includegraphics{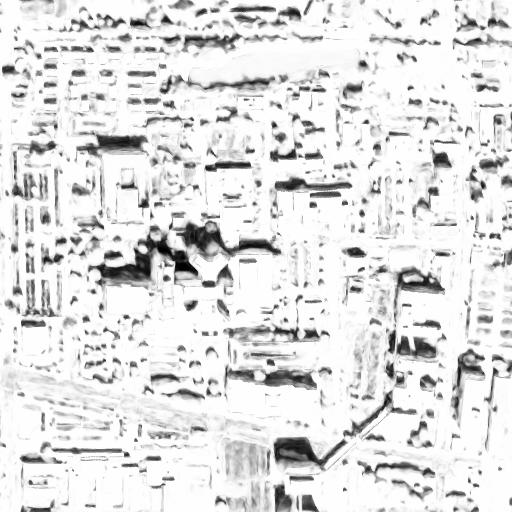}
    \\ \rule{0pt}{27pt}
    \rotatebox[origin=c]{90}{$\textbf{a}_\text{sky}$}
    & \includegraphics{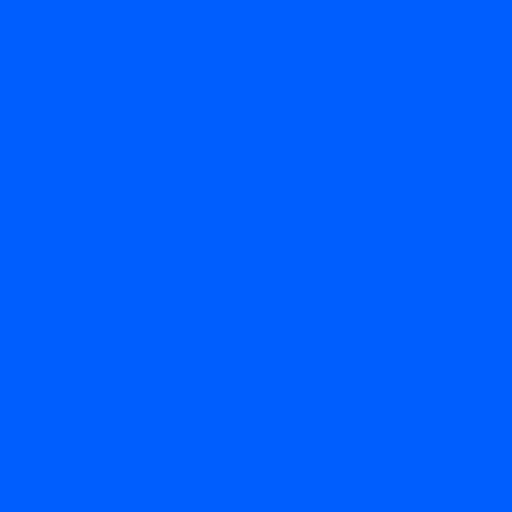}
    & \includegraphics{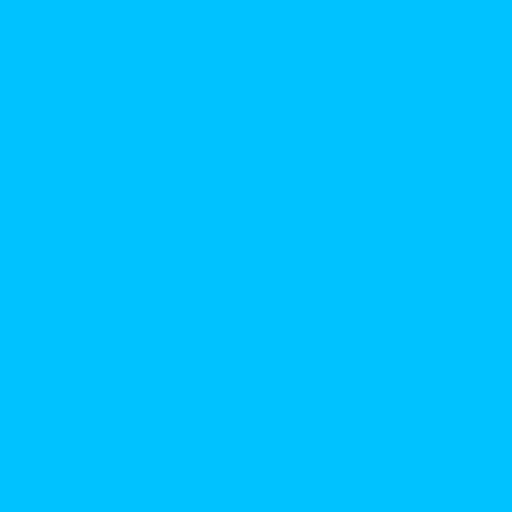}
    & \includegraphics{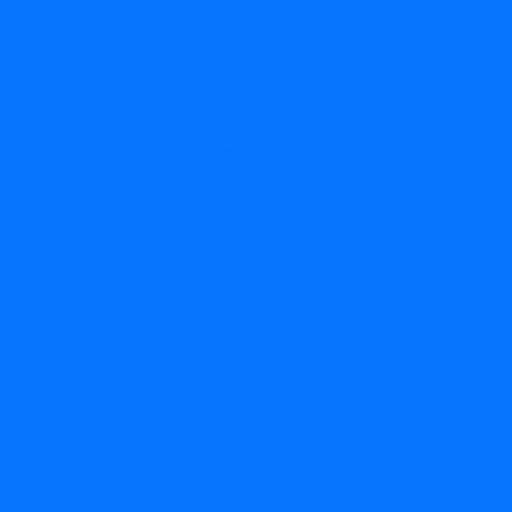}
    & \includegraphics{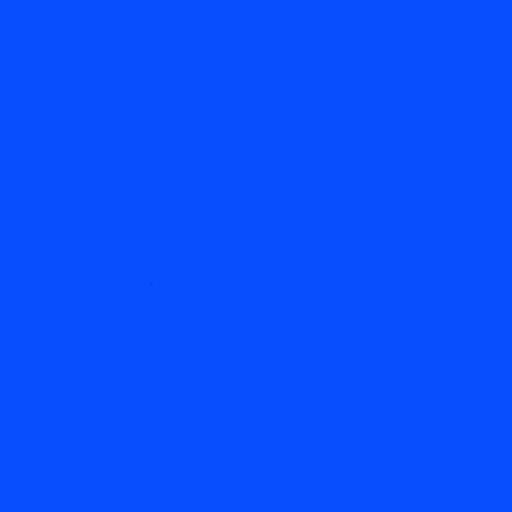}
  \end{tabularx}
\caption{Ground truth $\textbf{c}_{\text{gt}}$, shading scalar $s$, and ambient sky color predictions $\textbf{a}_\text{sky}$ for Sat-NeRF on the different Jacksonville areas (for two different runs).}
\label{fig:sat-nerf-sun-impact-jax}
\end{figure}

In contrast, this phenomenon is less pronounced when examining Figure \ref{fig:sat-nerf-sun-impact-jax}. In the images for JAX\_004, there are trends of the shading scalar $s$ having lower values for grass and trees than expected. Despite minimal seasonal variation in Jacksonville, differences in grass and tree color are still observable. For JAX\_068, where there are few trees and grass, the shading scalar consistently reaches low values in shaded areas.

When examining shading scalars and ambient sky colors in Jacksonville, it is apparent that the sky color remains consistently blue, despite predominant green or grey tones in the images. This suggests that the ambient color accurately predicts the sky color, in contrast with the situation in Omaha, see Figure \ref{fig:sat-nerf-sun-impact-oma}.

\subsubsection{Novel equinox sun direction}
\begin{figure*}
\centering
\centering
\setlength\tabcolsep{2pt}
\adjustboxset{width=\linewidth,valign=c}
      \begin{tabularx}{\linewidth}{@{}
      l
      X @{\hspace{1pt}}
      X @{\hspace{1pt}}
      X @{\hspace{1pt}}
      X @{\hspace{4pt}}
      X @{\hspace{1pt}}
      X @{\hspace{1pt}}
      X @{\hspace{1pt}}
      X @{\hspace{4pt}}
      X @{\hspace{1pt}}
      X @{\hspace{1pt}}
      X @{\hspace{1pt}}
      X @{\hspace{1pt}}
      X 
    @{}}
    & \multicolumn{4}{c}{\textbf{January}}
    & \multicolumn{4}{c}{\textbf{March}}
    & \multicolumn{4}{c}{\textbf{September}}
    \\ \cmidrule(lr{8pt}){2-5} \cmidrule(lr{8pt}){6-9} \cmidrule(lr{8pt}){10-13}
    
    & \multicolumn{1}{c}{\textbf{042}}
    & \multicolumn{1}{c}{\textbf{084}}
    & \multicolumn{1}{c}{\textbf{163}}
    & \multicolumn{1}{c}{\textbf{212}}
    & \multicolumn{1}{c}{\textbf{042}}
    & \multicolumn{1}{c}{\textbf{084}}
    & \multicolumn{1}{c}{\textbf{163}}
    & \multicolumn{1}{c}{\textbf{212}} 
    & \multicolumn{1}{c}{\textbf{042}}
    & \multicolumn{1}{c}{\textbf{084}}
    & \multicolumn{1}{c}{\textbf{163}}
    & \multicolumn{1}{c}{\textbf{212}}
    
    \\ %
    \rotatebox[origin=c]{90}{$\textbf{c}_\text{gt}$}
    & \includegraphics{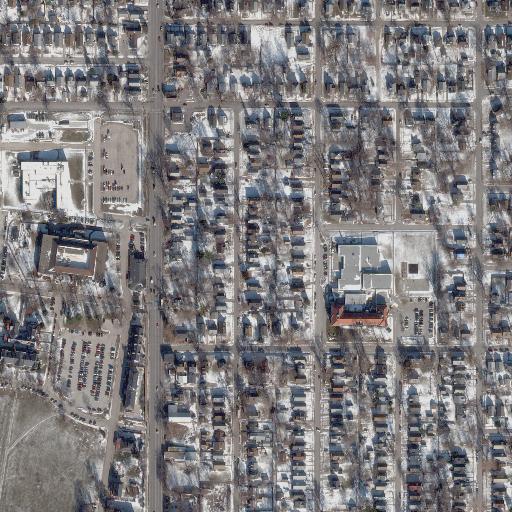}
    & \includegraphics{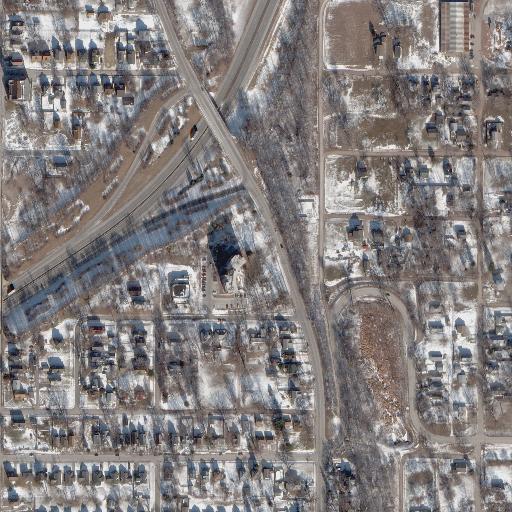}
    & \includegraphics{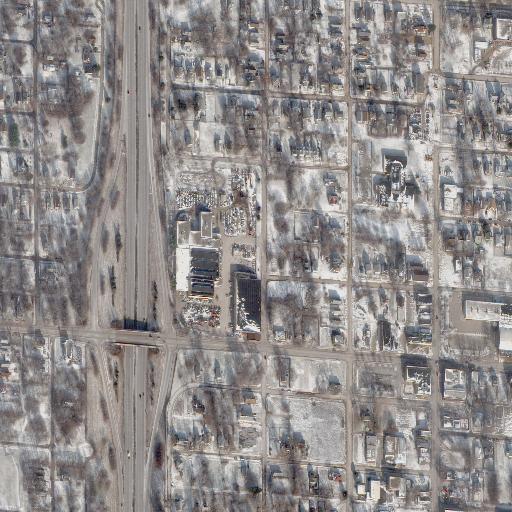}
    & \includegraphics{sec/img/gt_rgb/OMA_212_012_RGB_epoch20.jpg}
    & \includegraphics{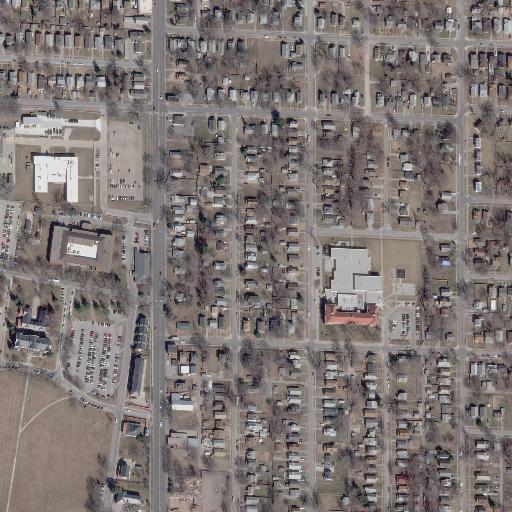}
    & \includegraphics{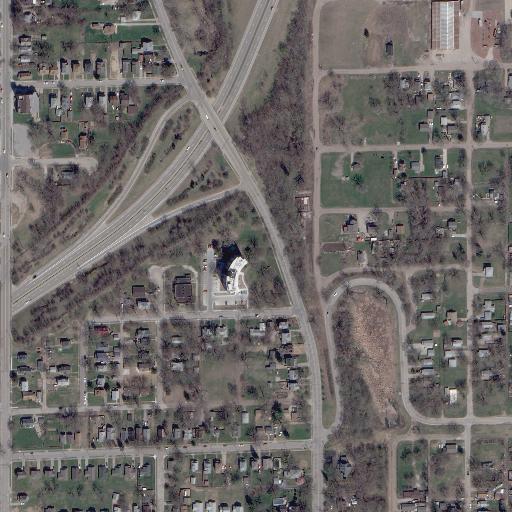}
    & \includegraphics{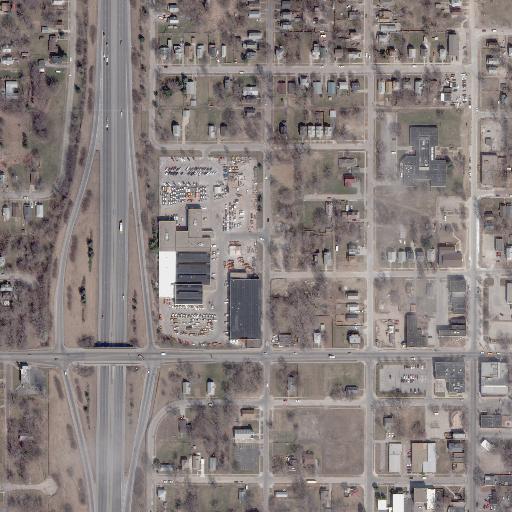}
    & \includegraphics{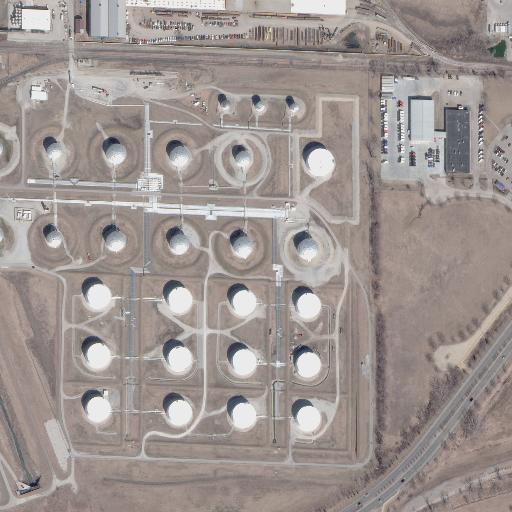}
    & \includegraphics{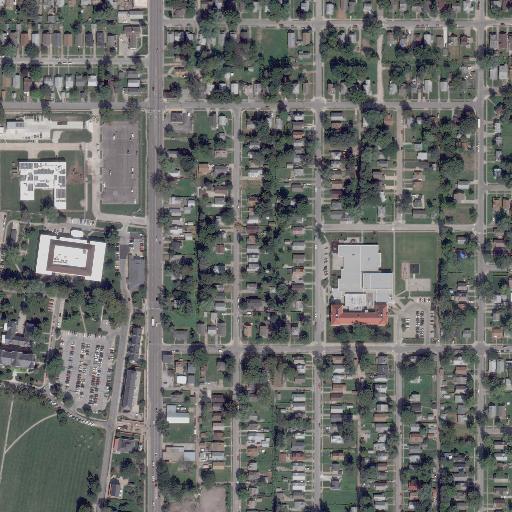}
    & \includegraphics{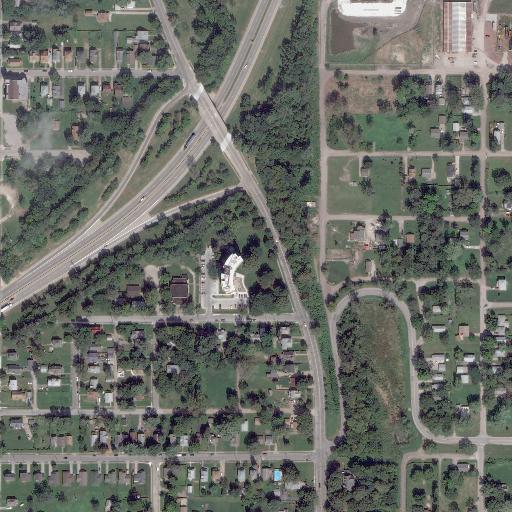}
    & \includegraphics{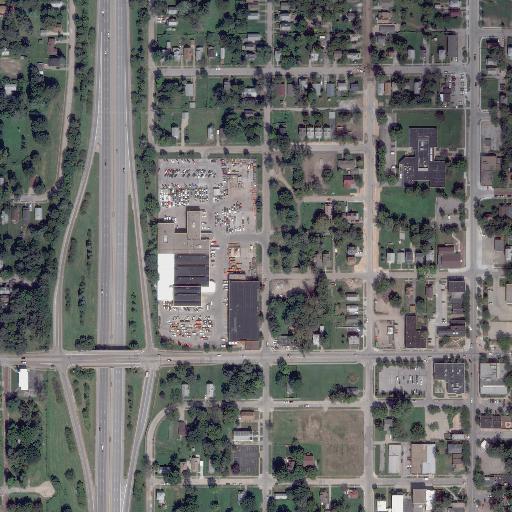}
    & \includegraphics{sec/img/gt_rgb/OMA_212_036_RGB_epoch20.jpg}
    \\ \rule{0pt}{16pt}
    \rotatebox[origin=c]{90}{SN} 
    & \includegraphics{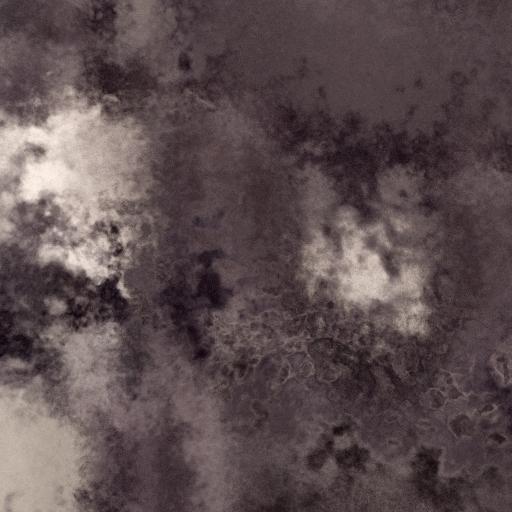}
    & \includegraphics{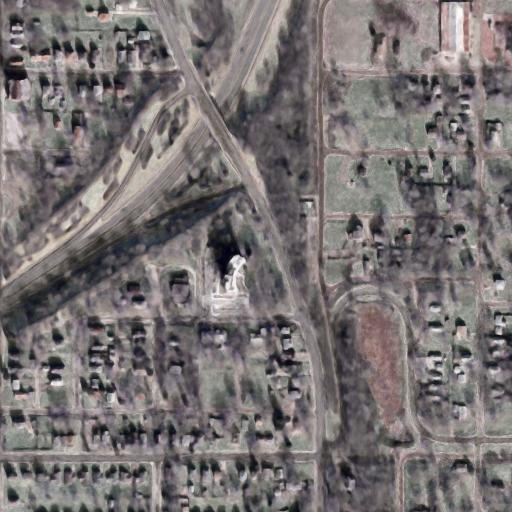}
    & \includegraphics{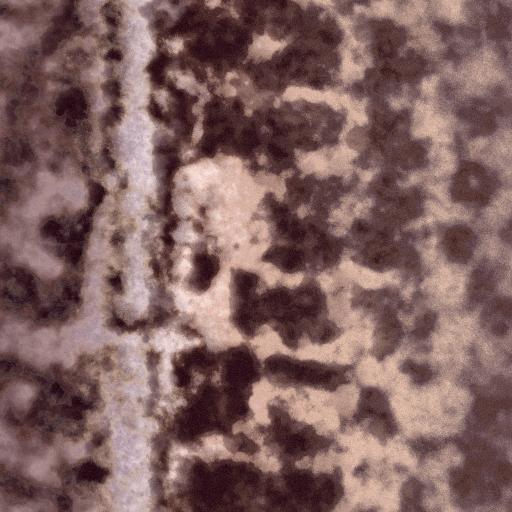}
    & \includegraphics{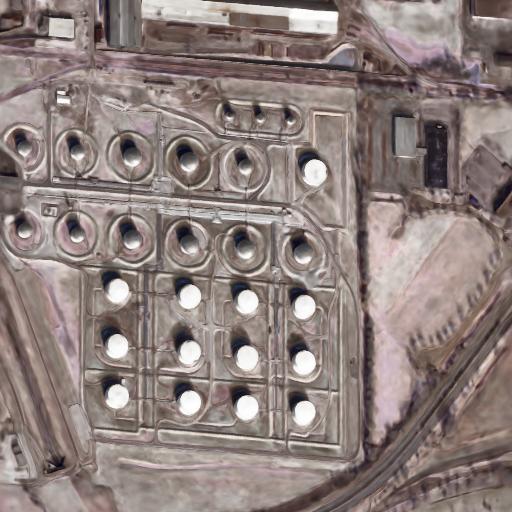}
    & \includegraphics{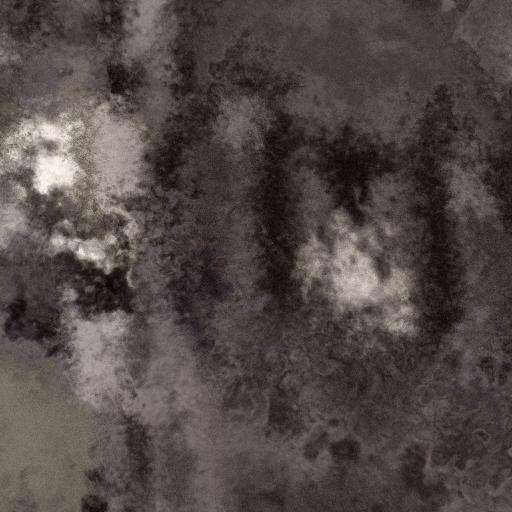}
    & \includegraphics{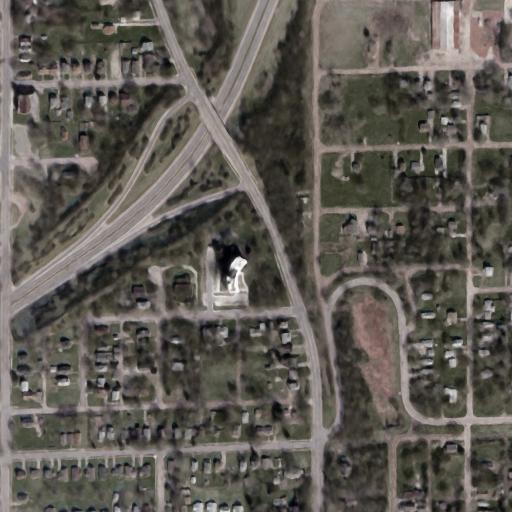}
    & \includegraphics{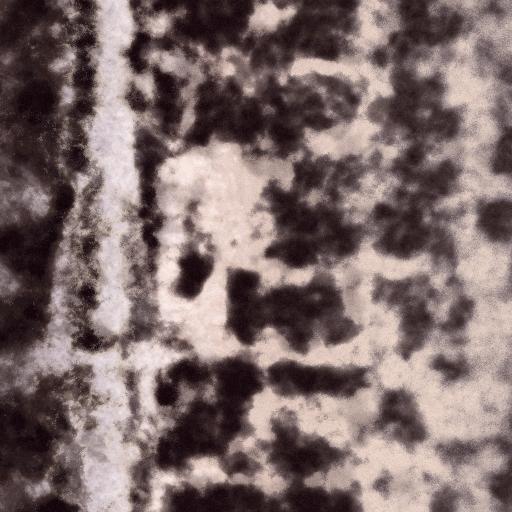}
    & \includegraphics{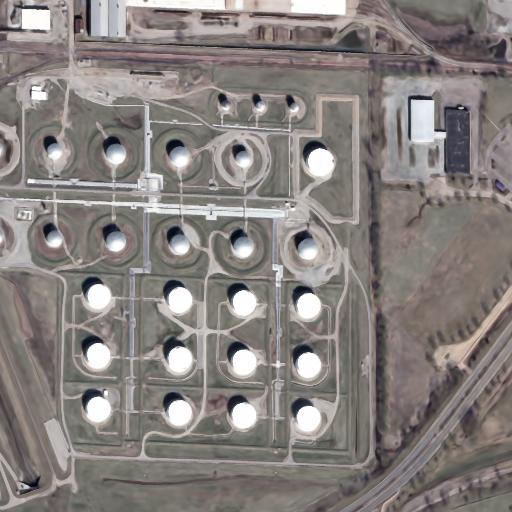}
    & \includegraphics{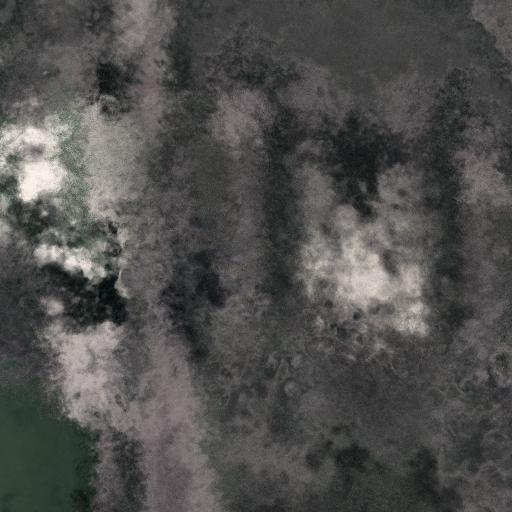}
    & \includegraphics{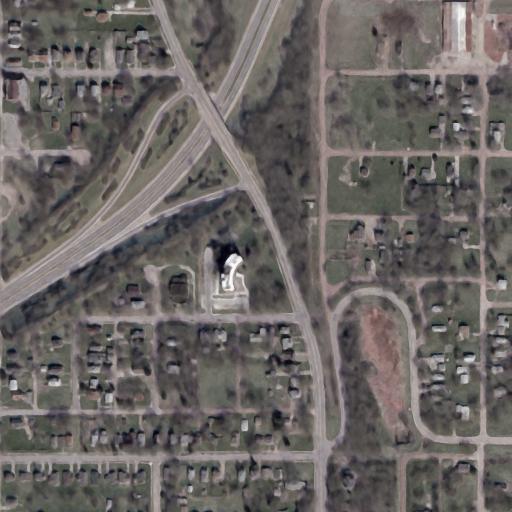}
    & \includegraphics{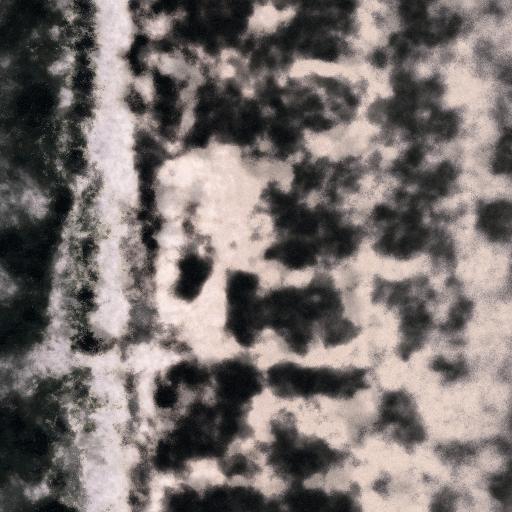}
    & \includegraphics{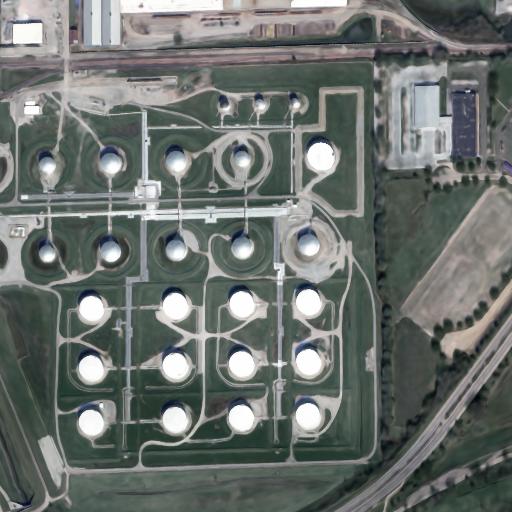}
    \\ \rule{0pt}{16pt}
    \rotatebox[origin=c]{90}{ME} 
    & \includegraphics{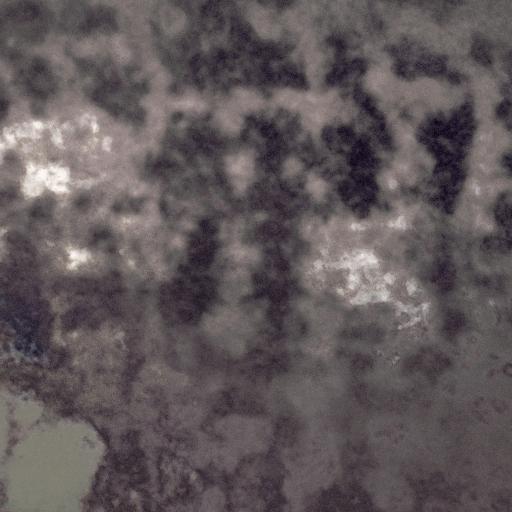}
    & \includegraphics{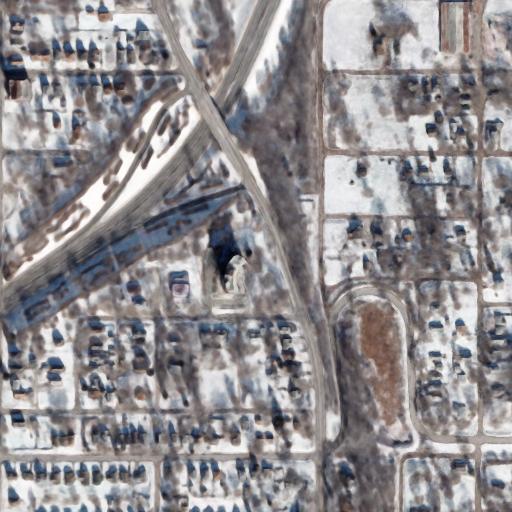}
    & \includegraphics{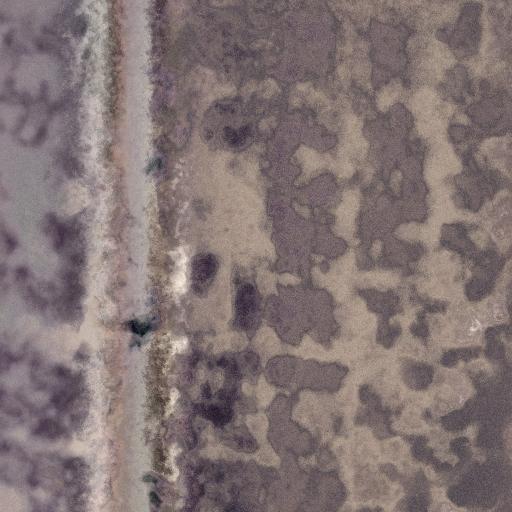}
    & \includegraphics{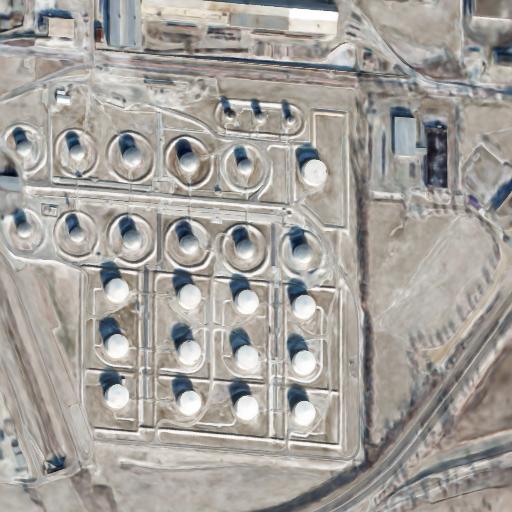}
    & \includegraphics{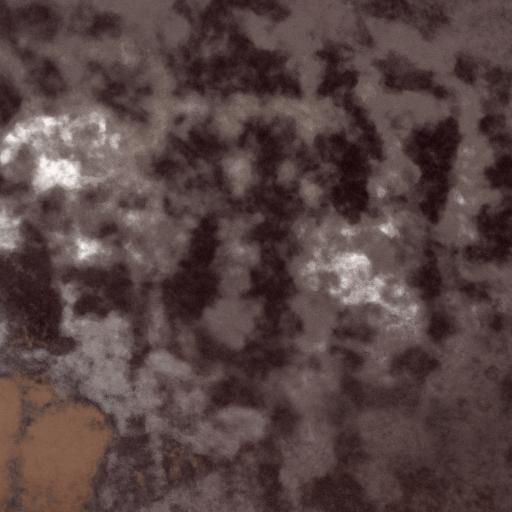}
    & \includegraphics{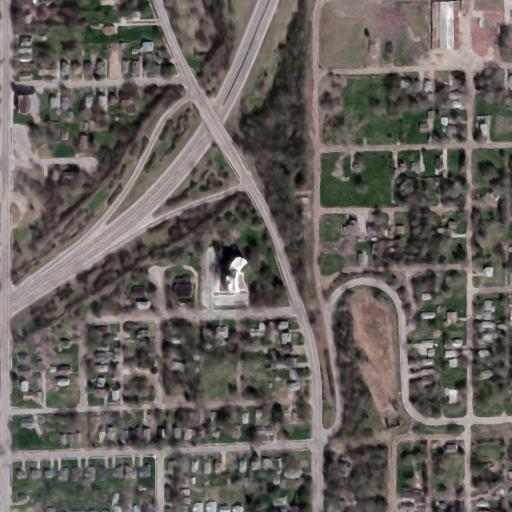}
    & \includegraphics{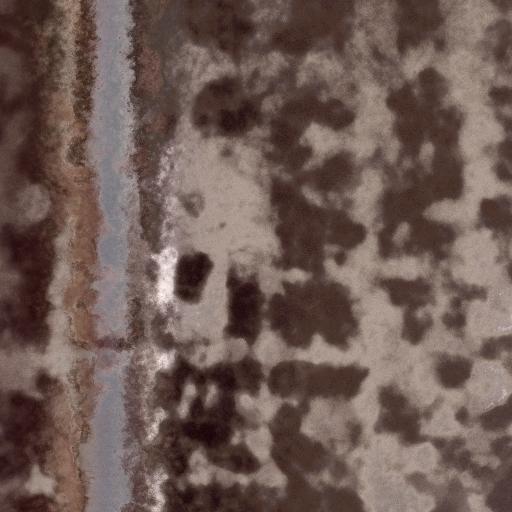}
    & \includegraphics{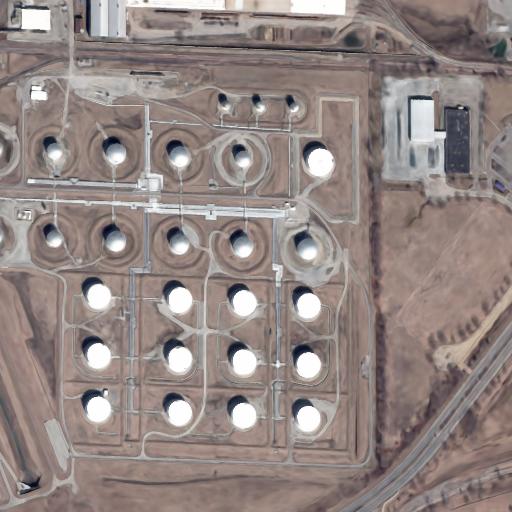}
    & \includegraphics{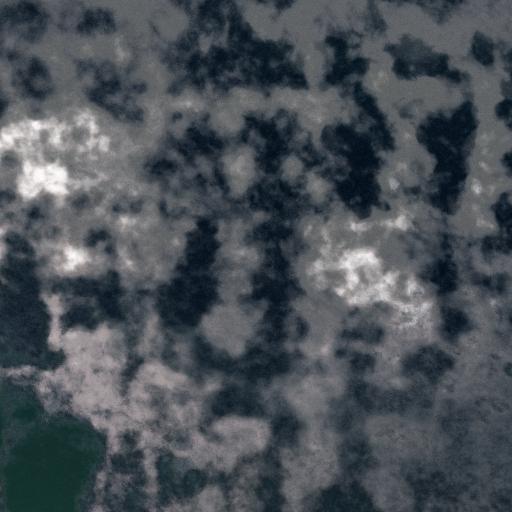}
    & \includegraphics{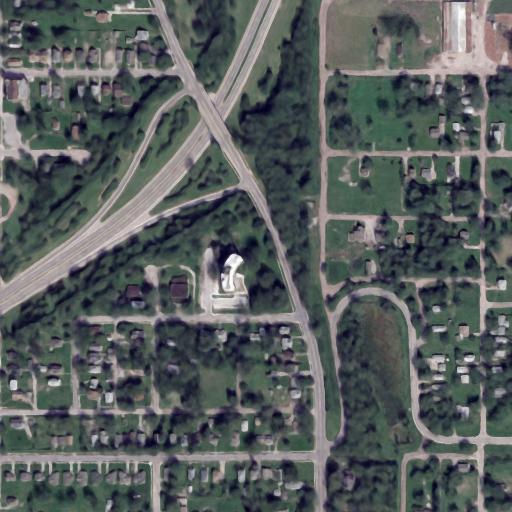}
    & \includegraphics{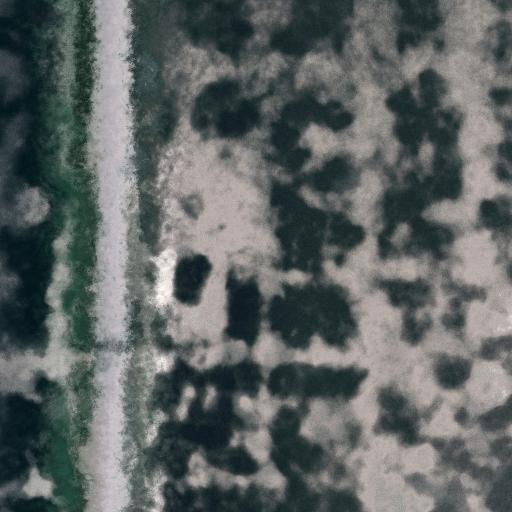}
    & \includegraphics{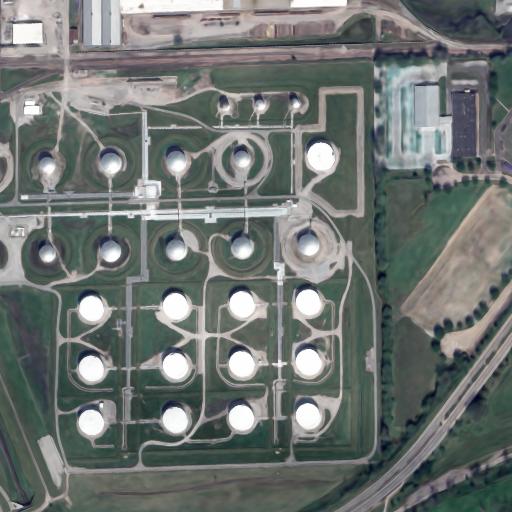}
    \\ \rule{0pt}{16pt}
    \rotatebox[origin=c]{90}{PE} 
    & \includegraphics{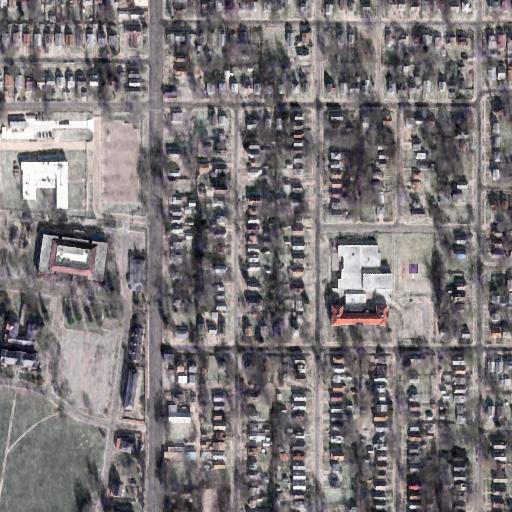}
    & \includegraphics{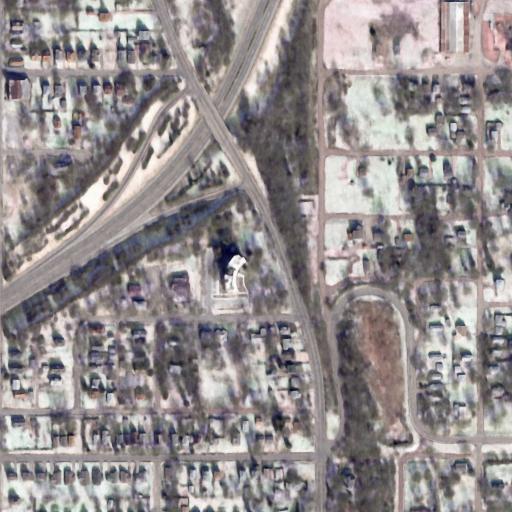}
    & \includegraphics{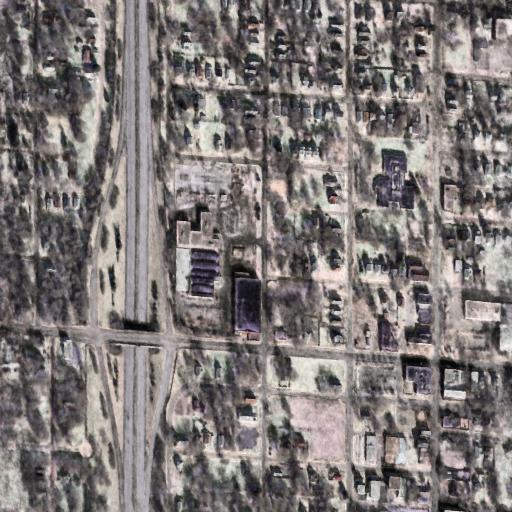}
    & \includegraphics{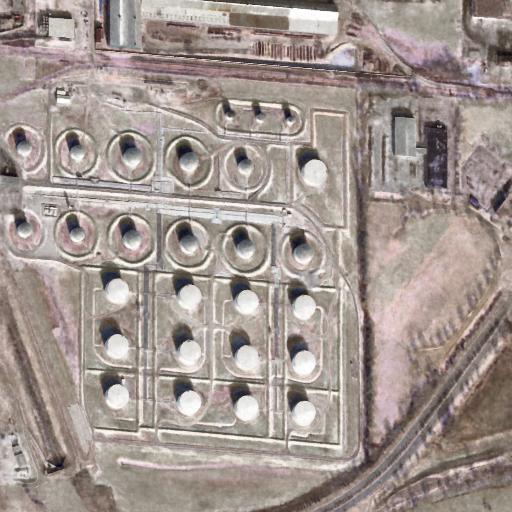}
    & \includegraphics{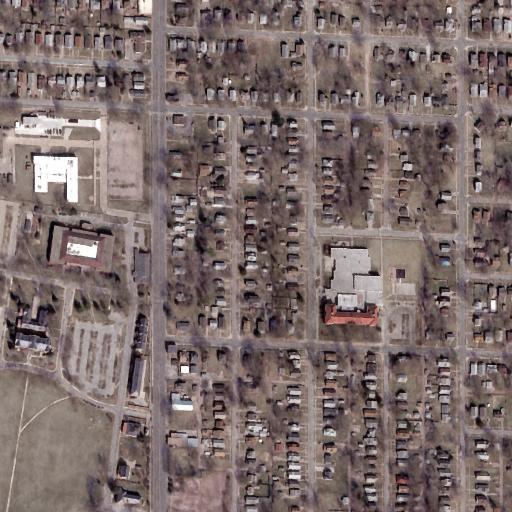}
    & \includegraphics{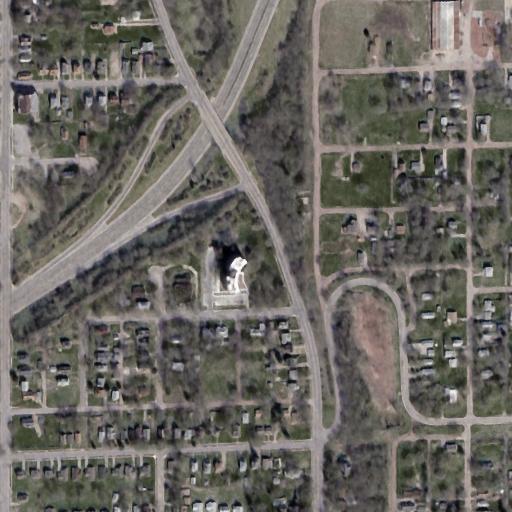}
    & \includegraphics{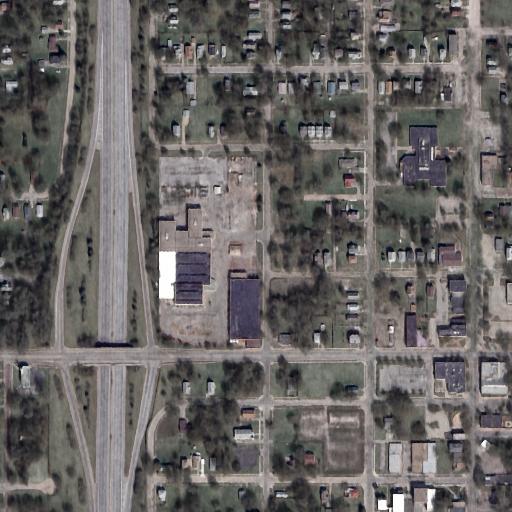}
    & \includegraphics{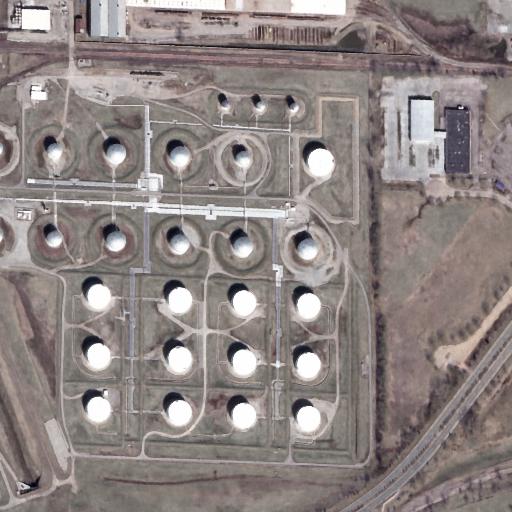}
    & \includegraphics{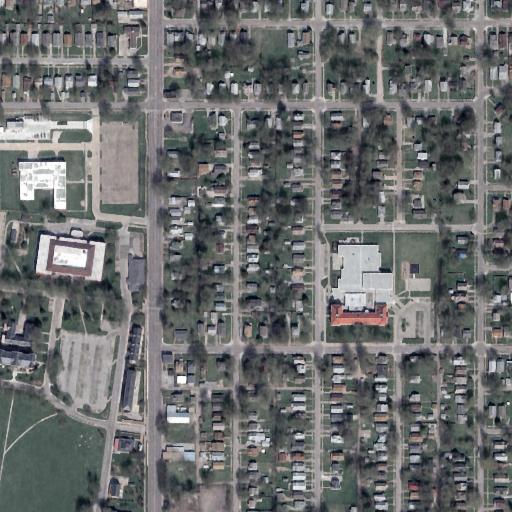}
    & \includegraphics{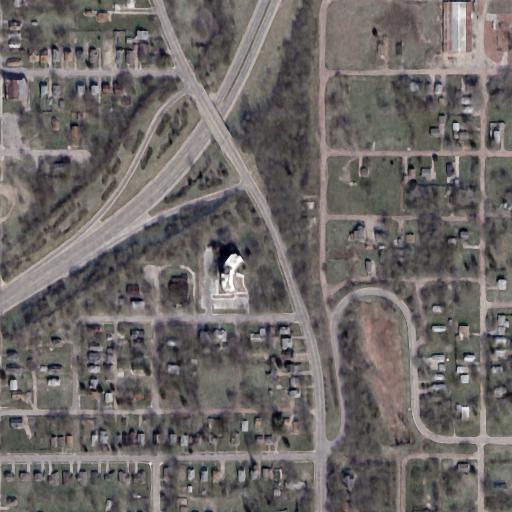}
    & \includegraphics{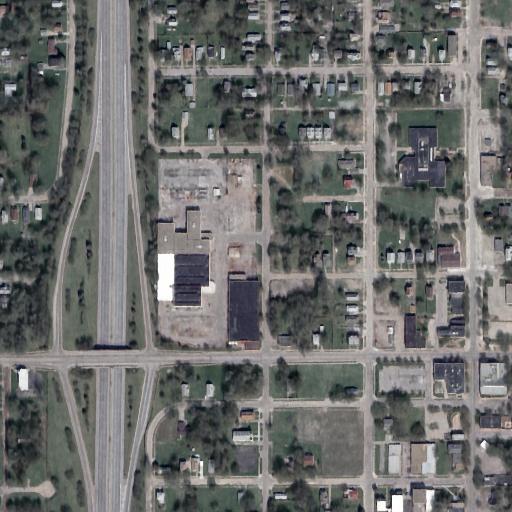}
    & \includegraphics{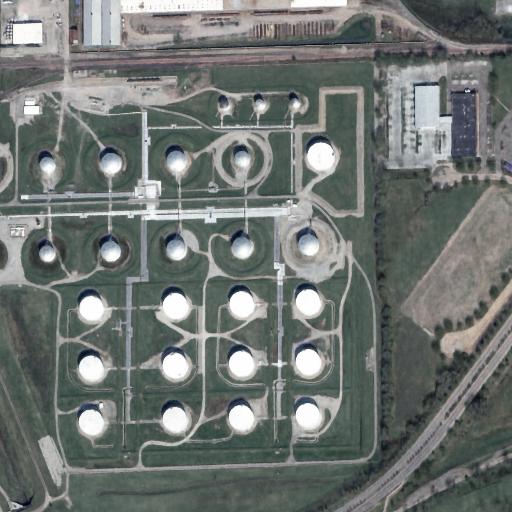}
    \\ \rule{0pt}{16pt}
    \rotatebox[origin=c]{90}{PN} 
    & \includegraphics{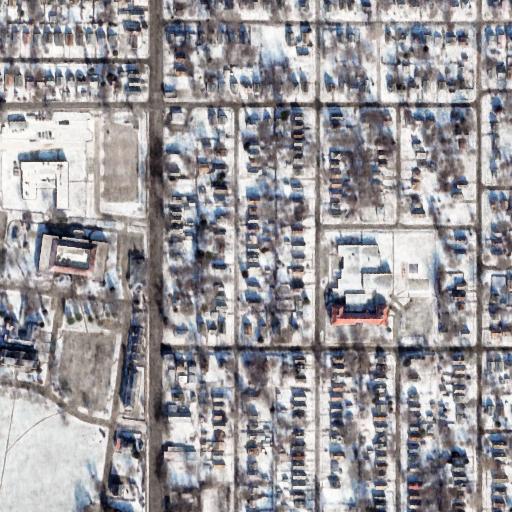}
    & \includegraphics{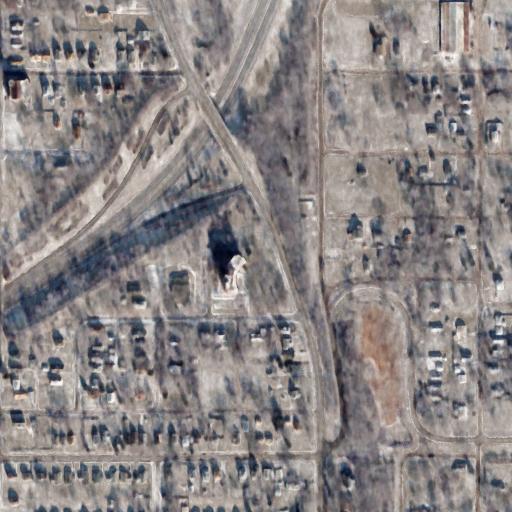}
    & \includegraphics{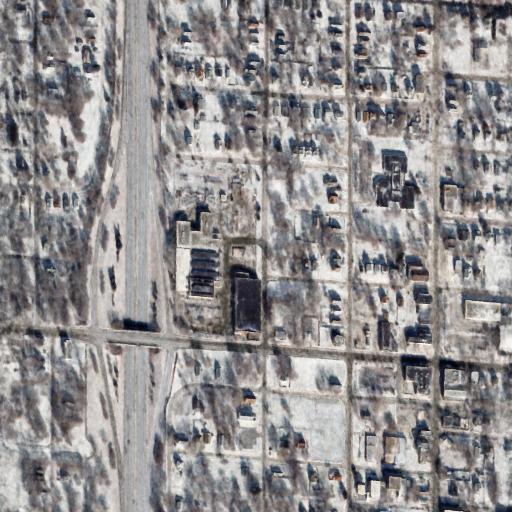}
    & \includegraphics{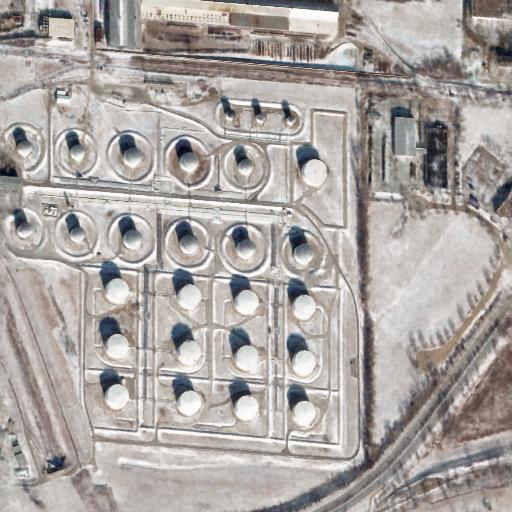}
    & \includegraphics{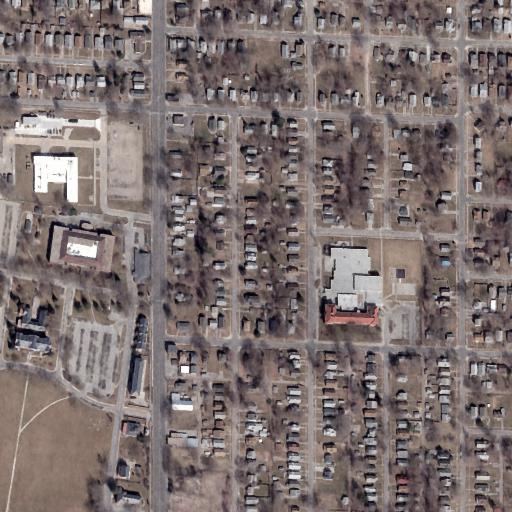}
    & \includegraphics{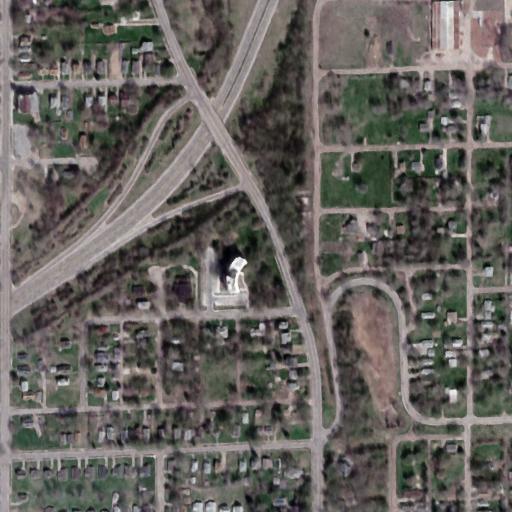}
    & \includegraphics{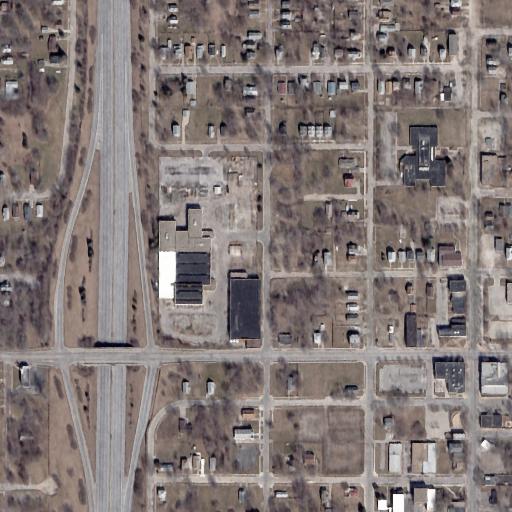}
    & \includegraphics{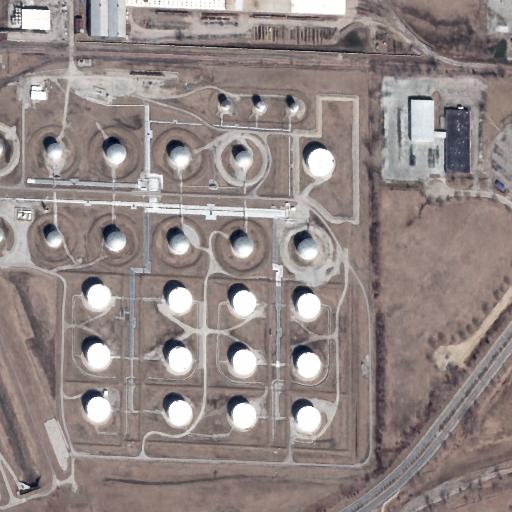}
    & \includegraphics{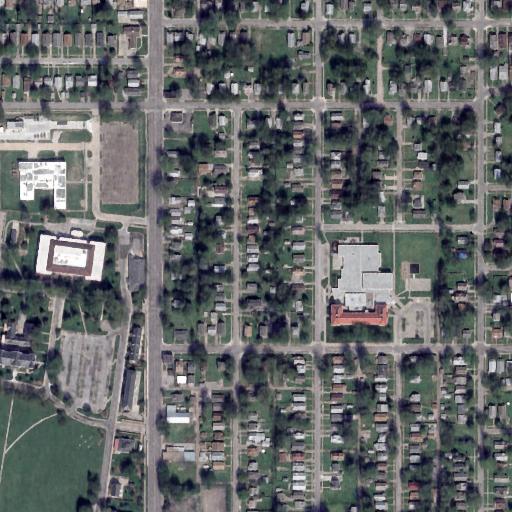}
    & \includegraphics{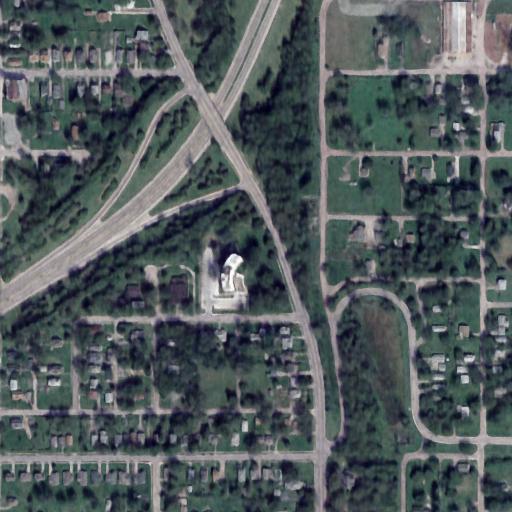}
    & \includegraphics{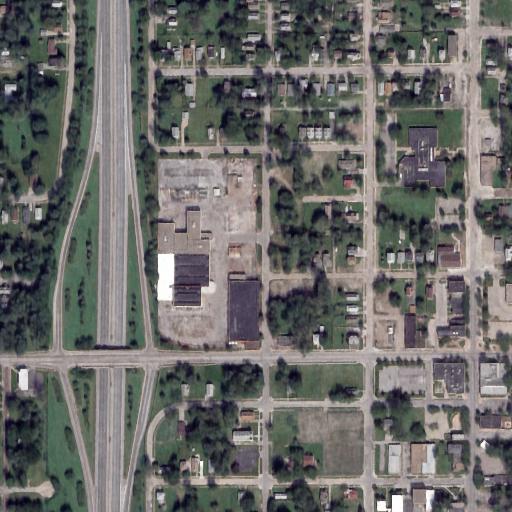}
    & \includegraphics{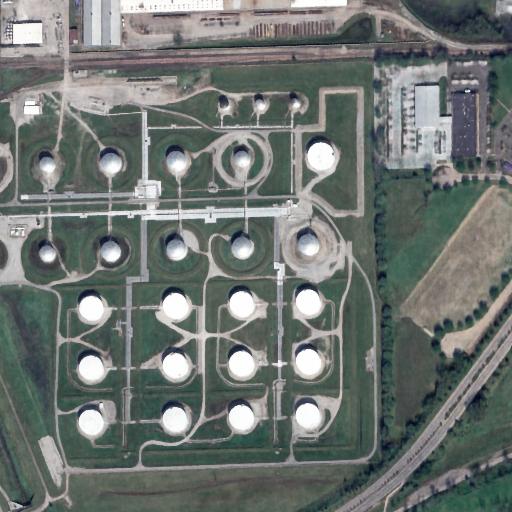}
  \end{tabularx}
\caption{RGB predictions for the Omaha areas for three different months representing three different seasons, using sun directions $\textbf{d}_{sun}$ corresponding to the 15'th at 17:00 UTC each month. OMA\_084 has too few images from March, hence, images from April were used instead. Abbreviations: $\textbf{c}_{\text{gt}}$: ground truth, SN: Sat-NeRF, ME: Sat-NeRF + month embedding, PE: Sat-NeRF + positional encoding, PN: Planet-NeRF (Sat-NeRF + month embedding + positional encoding).}
\label{fig:rgb_predictions}
\end{figure*}

In order to conduct a more in-depth analysis of Sat-NeRFs ability to discern distinctive features during equinox months, i.e. March and September, novel sun directions from these months were generated and used as input to the model to generate images with those specific sun directions. 


\begin{figure}[h]
  \adjustboxset{width=\linewidth,valign=c}
  \centering
  \begin{tabularx}{0.9\linewidth}{@{}
      l
      X @{\hspace{1pt}}
      X @{\hspace{1pt}}
      X @{\hspace{4pt}}
      X @{\hspace{1pt}}
      X @{\hspace{1pt}}
      X @{\hspace{1pt}}
      X 
    @{}}
    & \multicolumn{3}{c}{\textbf{March}} & \multicolumn{3}{c}{\textbf{September}} \\ \cmidrule(lr{8pt}){2-4} \cmidrule(lr{8pt}){5-7}
    & \multicolumn{1}{c}{\textbf{1st}}
    & \multicolumn{1}{c}{\textbf{15th}}
    & \multicolumn{1}{c}{\textbf{30th}}
    & \multicolumn{1}{c}{\textbf{1st}}
    & \multicolumn{1}{c}{\textbf{15th}}
    & \multicolumn{1}{c}{\textbf{30th}}
    \\
    \rotatebox[origin=c]{90}{\textbf{132}}
    & \includegraphics{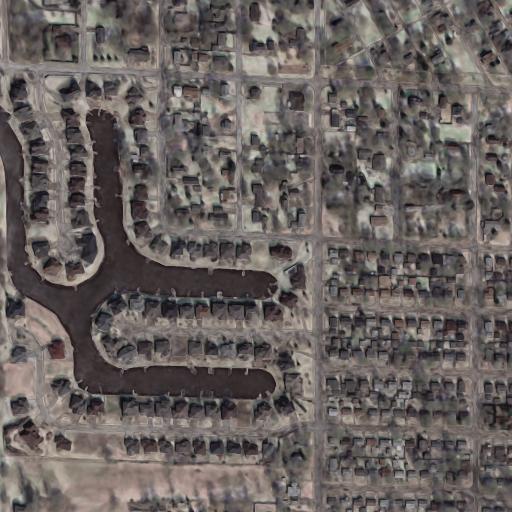}
    & \includegraphics{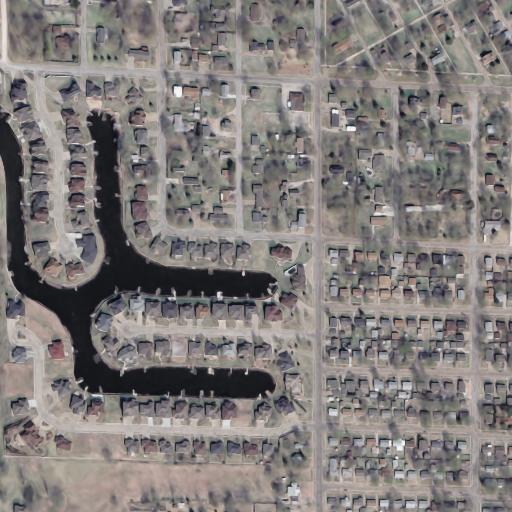}
    & \includegraphics{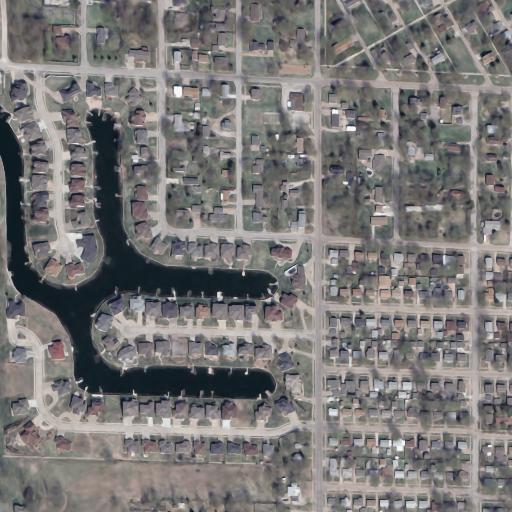}
    & \includegraphics{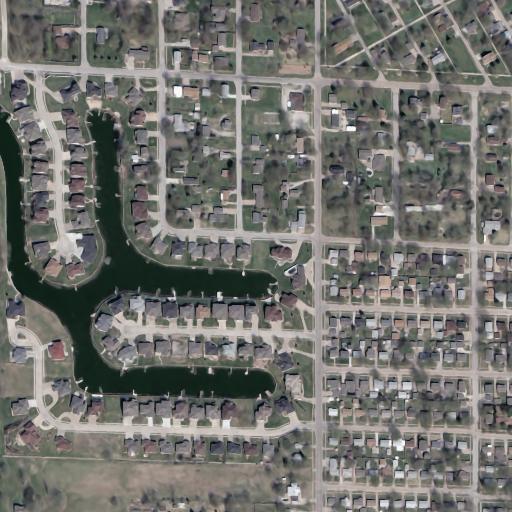}
    & \includegraphics{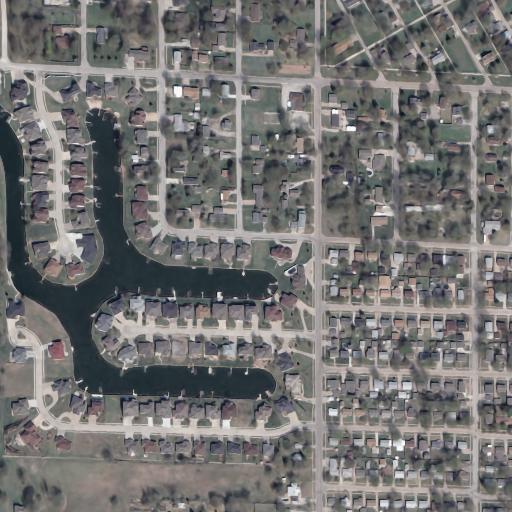}
    & \includegraphics{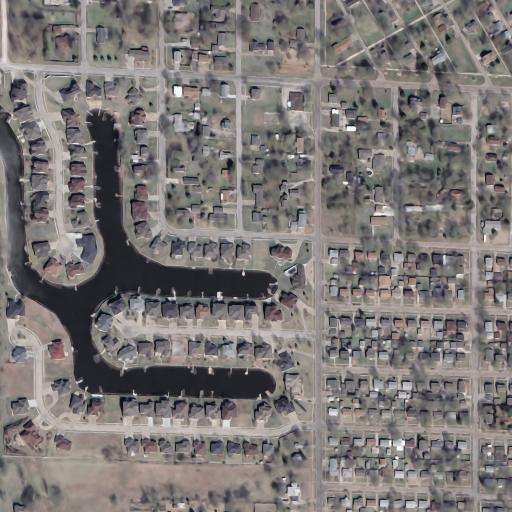}
    \\ \rule{0pt}{28pt}
   \rotatebox[origin=c]{90}{\textbf{212}}
    & \includegraphics{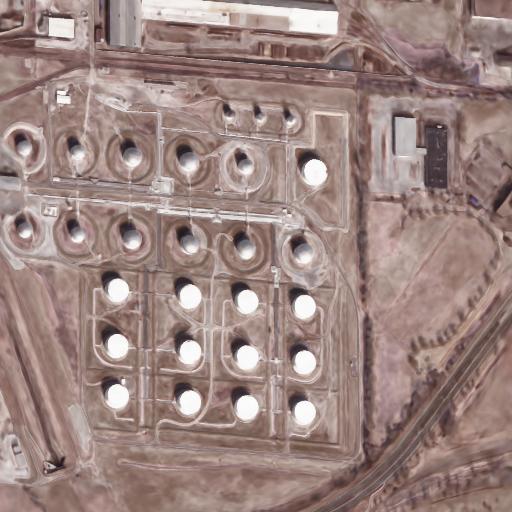}
    & \includegraphics{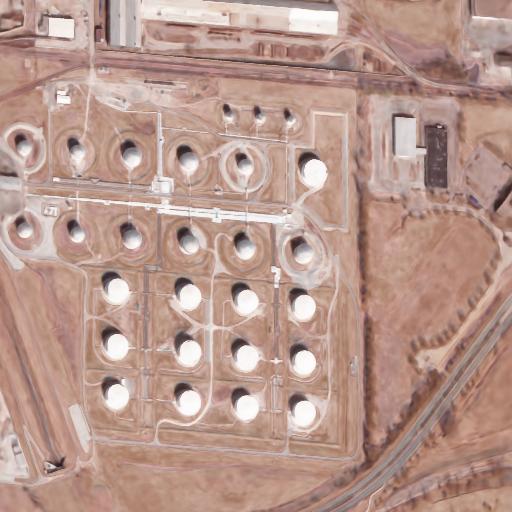}
    & \includegraphics{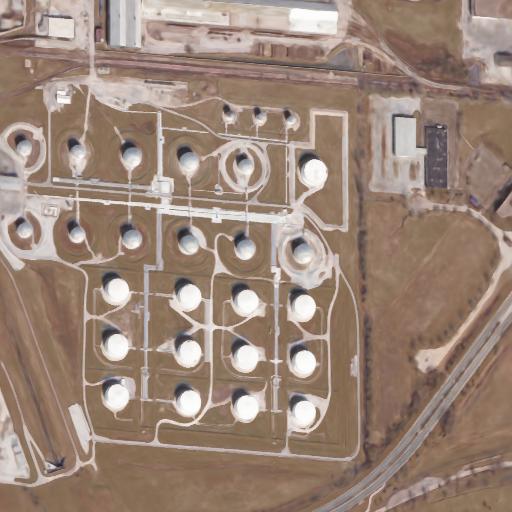}
    & \includegraphics{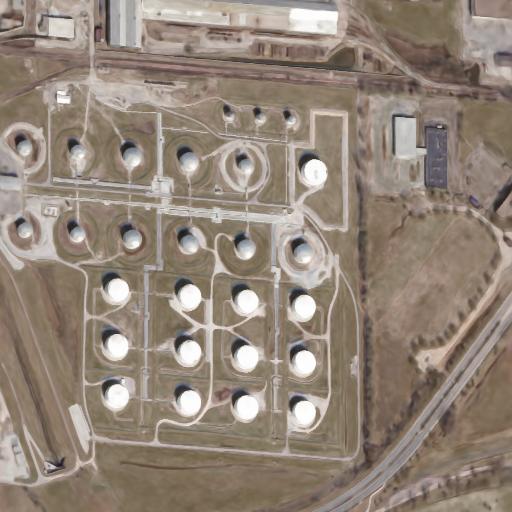}
    & \includegraphics{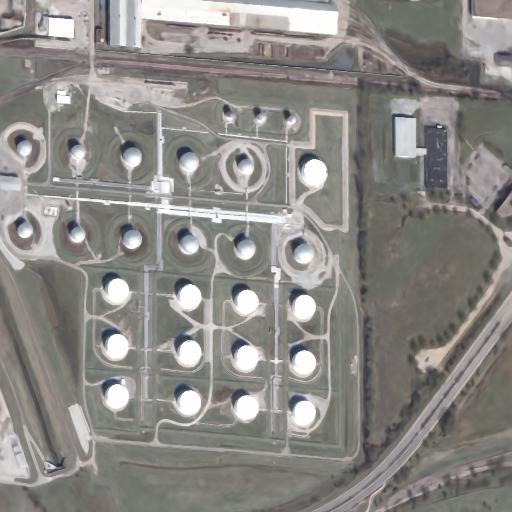}
    & \includegraphics{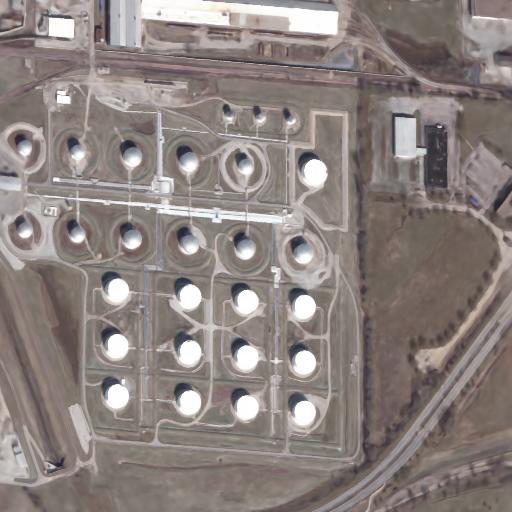}
    \\ \rule{0pt}{28pt}
    \rotatebox[origin=c]{90}{\textbf{374}}
    & \includegraphics{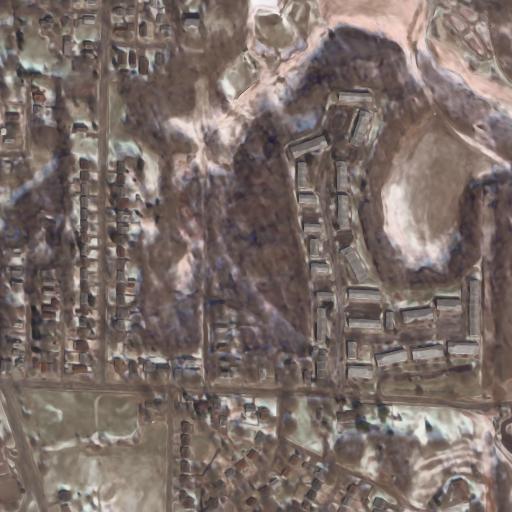}
    & \includegraphics{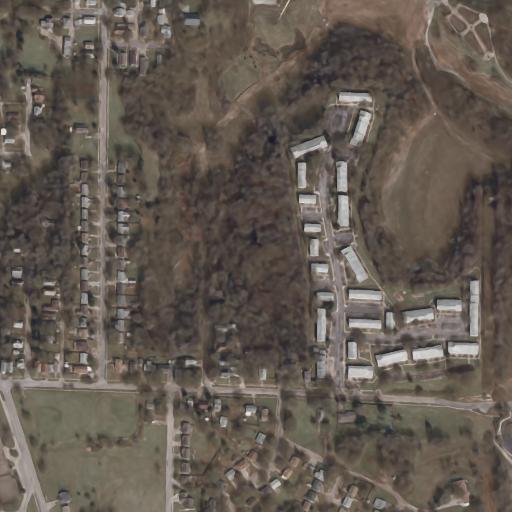}
    & \includegraphics{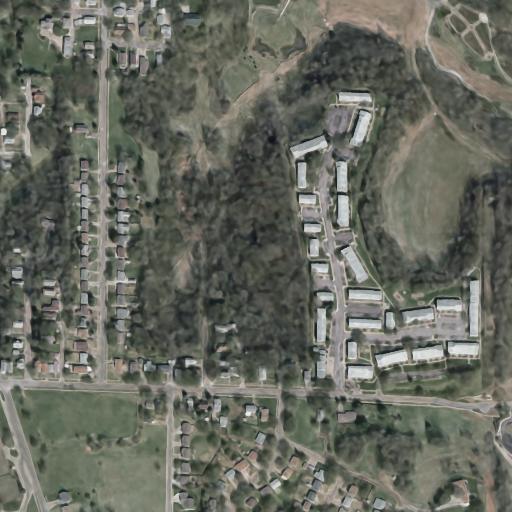}
    & \includegraphics{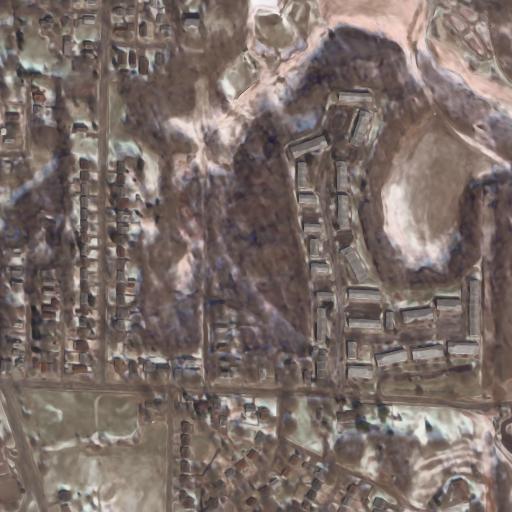}
    & \includegraphics{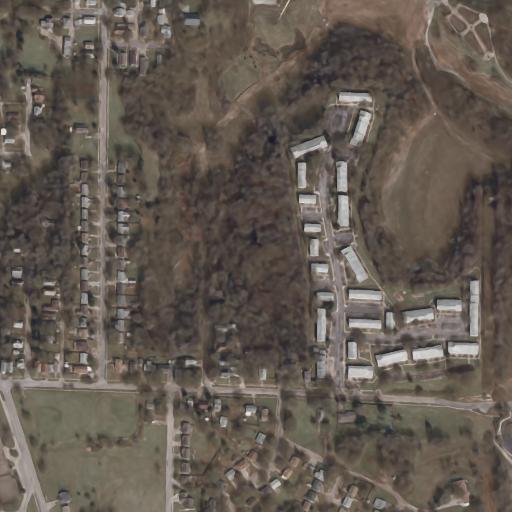}
    & \includegraphics{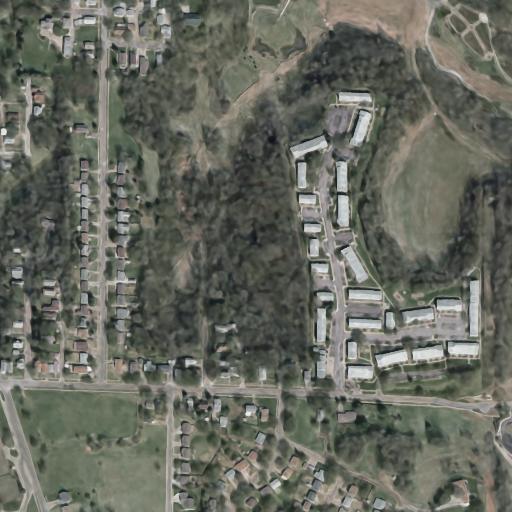}
  \end{tabularx}
\caption{RGB predictions by the Sat-NeRF model of three Omaha areas demonstrating how Sat-NeRF confuses the equinox months March and September.}
\label{img:pred-mars-sep}
\end{figure}

Sun direction vectors $\textbf{d}_{sun}$ were generated for the 1st, 15th, and 30th of March and September. The RGB predictions for Sat-NeRF, depicted in Figure \ref{img:pred-mars-sep}, reveal a noteworthy trend where the model confuses March and September. September appears overly brown across all datasets, while March is mostly accurate but occasionally shows excessive green in the last image from OMA\_374, which is not representative of that time of year (compare with $\textbf{c}_{\text{gt}}$ in Figure \ref{fig:rgb_predictions}). Examining the images from OMA\_374, they are nearly identical. These findings reveal a limitation in the model's ability to represent the entire year accurately. While it learns from solar path details, it struggles with nuanced variations during equinox months. The confusion between March and September highlights the complexity of accurately modeling seasonal transitions, highlighting the need for further refinement and considerations to enhance the model's overall performance.



\if{false}
\begin{table*}
\centering
\begin{tabular}{l|l l l l|l l l l|l l l l}
\hline
\textbf{Model}& \multicolumn{4}{c|}{\textbf{PSNR} $\uparrow$} & \multicolumn{4}{c|}{\textbf{SSIM} $\uparrow$} & \multicolumn{4}{c}{\textbf{Alt. MAE [m]} $\downarrow$}  \\ [0.5ex] 

AOI: OMA &  212 & 132 & 374 & Mean & 212 & 132 & 374 & 
Mean & 212 & 132 & 374 & Mean \\
\hline \hline
Sat-NeRF &  21.87 & 19.87 & 18.04& 19.93 & 0.74 & 0.71 & 0.62 & 0.69 & 0.84  & 1.15 & \textbf{4.23}  & 2.07\\
Season-NeRF  & 17.79 & 19.82  & \textbf{20.48}  &19.36 & 0.57 &0.62 &0.52 & 0.57 & 0.94 & 1.22 & 4.44 & 2.2 \\
Planet-NeRF &  \textbf{22.40}  & \textbf{21.15}  & 20.29 & \textbf{21.28} & 0.76  & \textbf{0.78}  & 0.70  & \textbf{0.75} & \textbf{0.82}  &1.12  & 4.25  & \textbf{2.06} \\
\hline
\end{tabular}
\caption{PSNR, SSIM, and altitude MAE across all Omaha areas and various model types. The value reported for Season-NeRF was sourced from their paper \cite{season-nerf}, specifically referencing the results from their full model.}
\label{table:results}
\end{table*}
\fi

\if{false} 
\begin{table*}
\centering
\caption{SSIM and altitude MAE across all Omaha areas and various model types. SN: Sat-NeRF, ME: Sat-NeRF + month embedding, PE: Sat-NeRF + positional encoding, SE: Season-NeRF, PN: Planet-NeRF, SDF: Planet-NeRF + SDF}
\begin{tabular}{c|l l l l l l |l l l l l l }
\hline
\textbf{AOI:OMA}& \multicolumn{6}{c|}{\textbf{SSIM} $\uparrow$} & \multicolumn{6}{c}{\textbf{Alt. MAE [m]} $\downarrow$}  \\ [0.5ex] 

Model & SN & ME & PE & SE & PN & SDF & SN & ME & PE & SE & PN & SDF \\
\hline \hline
042 & 0.25 & 0.30 & 0.69 & 0.60 & \textbf{0.75} & 0.70 & 3.40 & 3.36 & 1.94 & 2.47 & \textbf{1.89} &  2.10 \\
084 & 0.62 & \textbf{0.71} & 0.67 & 0.55 & 0.67 & 0.64 & 3.34 & 2.95 & 3.24 & 2.82 & \textbf{2.58} &  3.07 \\
132 & 0.71 & \textbf{0.78} & 0.69 & 0.62 & \textbf{0.78} & 0.73 & 1.15 & 1.12 & 1.34 & 1.22 & \textbf{1.05} & 1.10 \\
163 & 0.38 & 0.42 & 0.72 & 0.52 & \textbf{0.76} & \textbf{0.76} & 3.06 & 3.20 & \textbf{1.40} & 1.43 & 1.44 & 1.60 \\
203 & 0.39 & 0.45 & 0.69 & 0.65 & \textbf{0.72} & 0.70 &  3.78 & 4.54 & \textbf{1.16} & 1.17 & 1.20 &  1.38 \\
212 & 0.74 & 0.76 & \textbf{0.77} & 0.57 & \textbf{0.77} & \textbf{0.77} & 0.84  & \textbf{0.82} & 0.88 & 0.94 & 0.85 & 0.91 \\
281 & 0.43 & 0.40 & 0.71 & 0.60 & \textbf{0.75} & 0.72 & 4.61 & 4.86 & 1.60 & \textbf{1.46} & 1.58 & 1.51 \\
374 & 0.62 & 0.70 & 0.58 & 0.52 & \textbf{0.72} & 0.71 & 4.23 & 4.25 & \textbf{3.88} & 4.44 & 3.96 & 4.51 \\
\hdashline
Mean & 0.52 & 0.56 & 0.69 & 0.58 & \textbf{0.74} & 0.72 & 3.05 & 3.14 & 1.93 & 1.99 & \textbf{1.82} &  2.02 \\

\hline
\end{tabular}
\label{table:results_ssim_mae}
\end{table*}
\fi

\iftrue
\begin{table*}
\centering
\caption{PSNR, SSIM, and altitude MAE across all Omaha areas and various model types. SN: Sat-NeRF, ME: Sat-NeRF + month embedding, PE: Sat-NeRF + positional encoding, SE: Season-NeRF, PN: Planet-NeRF. The arrows indicate if higher $\uparrow$ or lower $\downarrow$ is better. Best results are marked in \textbf{bold}.}
\begin{tabular}{c|l l l l l |l l l l l |l l l l l }
\hline
 & \multicolumn{5}{c|}{\textbf{PSNR} $\uparrow$} & \multicolumn{5}{c|}{\textbf{SSIM} $\uparrow$} & \multicolumn{5}{c}{\textbf{Alt. MAE [m]} $\downarrow$}  \\ [0.5ex] 

OMA & SN & ME & PE & SE & PN & SN & ME & PE & SE & PN & SN & ME & PE & SE & PN \\
\hline \hline
042 & 15.0 & 15.4 & 19.3 & 19.4 & \textbf{20.1} & 0.25 & 0.30 & 0.69 & 0.60 & \textbf{0.75} & 3.40 & 3.36 & 1.94 & 2.47 & \textbf{1.89} \\
084 & 19.4 & 20.0 & 19.8 & 19.2 & \textbf{20.1} & 0.62 & \textbf{0.71} & 0.67 & 0.55 & 0.67 & 3.34 & 2.95 & 3.24 & 2.82 & \textbf{2.58} \\
132 & 19.9 & 21.2 & 19.8 & 19.8 & \textbf{21.4} & 0.71 & \textbf{0.78} & 0.69 & 0.62 & \textbf{0.78} & 1.15 & 1.12 & 1.34 & 1.22 & \textbf{1.05} \\
163 & 16.2 & 16.6 & 19.0 & 18.7 & \textbf{20.6} & 0.38 & 0.42 & 0.72 & 0.52 & \textbf{0.76} & 3.06 & 3.20 & \textbf{1.40} & 1.43 & 1.44 \\
203 & 16.7 & 16.9 & 19.6 & \textbf{21.0} & 20.2 & 0.39 & 0.45 & 0.69 & 0.65 & \textbf{0.72} & 3.78 & 4.54 & \textbf{1.16} & 1.17 & 1.20 \\
212 & 21.9 & 22.4 & \textbf{22.9} & 17.8 & 22.6 & 0.74 & 0.76 & \textbf{0.77} & 0.57 & \textbf{0.77} & 0.84  & \textbf{0.82} & 0.88 & 0.94 & 0.85 \\
281 & 16.4 & 16.3 & 19.6 & 19.7 & \textbf{20.0} & 0.43 & 0.40 & 0.71 & 0.60 & \textbf{0.75} & 4.61 & 4.86 & 1.60 & \textbf{1.46} & 1.58 \\
374 & 18.0 & 20.3 & 17.2 & \textbf{20.5} & \textbf{20.5} & 0.62 & 0.70 & 0.58 & 0.52 & \textbf{0.72} & 4.23 & 4.25 & \textbf{3.88} & 4.44 & 3.96 \\
\hdashline
Mean & 17.9 & 18.6 & 19.7 & 19.5 & \textbf{20.7} & 0.52 & 0.56 & 0.69 & 0.58 & \textbf{0.74} & 3.05 & 3.14 & 1.93 & 1.99 & \textbf{1.82} \\

\hline
\end{tabular}
\label{table:results_ssim_mae}
\end{table*}
\fi

\subsection{Planet-NeRF}
\label{sec:result-RQ3-planetnerf}
In this section, the results from the Sat-NeRF modification, Planet-NeRF, is presented and compared with the results obtained by Sat-NeRF \cite{sat-nerf} and Season-NeRF \cite{season-nerf}. Table \ref{table:results_ssim_mae} presents PSNR, SSIM, and altitude MAE scores for all models across all data sets and Figure \ref{fig:rgb_predictions} the RGB predictions for four of the eight Omaha areas. Planet-NeRF achieves top mean scores in PSNR and SSIM, with the highest mean performance in altitude MAE as well.




\subsubsection{Qualitative and quantitative results}
Upon reviewing the qualitative and quantitative results for Sat-NeRF+PE, it becomes evident that the inclusion of positional encoding enhances Sat-NeRF's effectiveness across all regions, with significant improvements in areas OMA\_042, OMA\_163, OMA\_203, and OMA\_281, which posed the largest challenges for Sat-NeRF and Sat-NeRF+ME. Figure \ref{fig:rgb_predictions} reveals that Sat-NeRF+PE yields clearer image predictions compared to both Sat-NeRF and Sat-NeRF+ME. However, it is apparent that, similar to Sat-NeRF, Sat-NeRF+PE still struggles to accurately capture seasonal variations, an aspect where Sat-NeRF+ME excels. When observing Figure \ref{fig:rgb_predictions} it is clear that Sat-NeRF+ME more accurately captures the seasonal variations for areas OMA\_042, OMA\_163, OMA\_203, and OMA\_281 compared to both Sat-NeRF and Sat-NeRF+PE.

Observing the results for Planet-NeRF, which integrates Sat-NeRF with both ME and PE, it is clear to see that the performance greatly increases. Planet-NeRF can, thanks to, positional encoding and the month embedding vectors truthfully represent the seasonal variations for all areas, which is clear both visually in Figure \ref{fig:rgb_predictions} as well as in Table \ref{table:results_ssim_mae}.

Furthermore, when inspecting the predicted ambient sky colors for Planet-NeRF in Figure \ref{fig:sat-nerf-sun-impact-jax}, Planet-NeRF predicts more accurate ambient sky colors than Sat-NeRF. In addition, Planet-NeRF do not, in contrast to Sat-NeRF get confused by the equinox months as can be seen in Figure \ref{fig:rgb_predictions}.

\subsubsection{Impact of sun direction}

\begin{figure}[h]
  \setlength\tabcolsep{2pt}
  \adjustboxset{width=1\linewidth,valign=c}
  \centering
  \begin{tabularx}{0.6\linewidth}{@{}
      X @{\hspace{1pt}}
      X @{\hspace{1pt}}
      X 
    @{}}
    \multicolumn{1}{c}{\textbf{Beginning}}
    & \multicolumn{1}{c}{\textbf{Middle}}
    & \multicolumn{1}{c}{\textbf{End}}
    \\
    \includegraphics{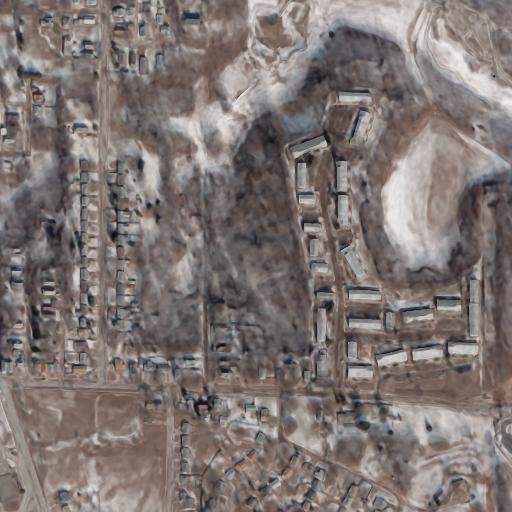}
    & \includegraphics{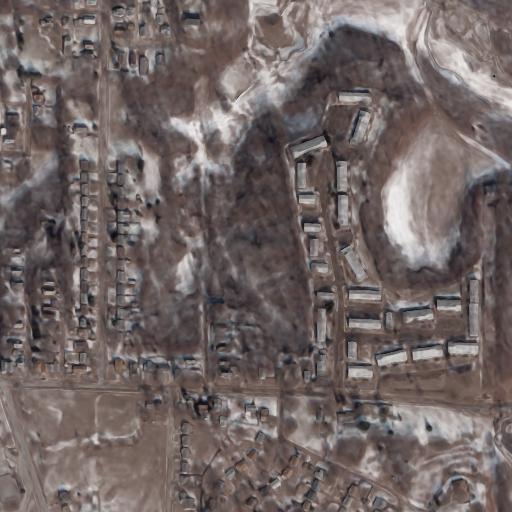}
    & \includegraphics{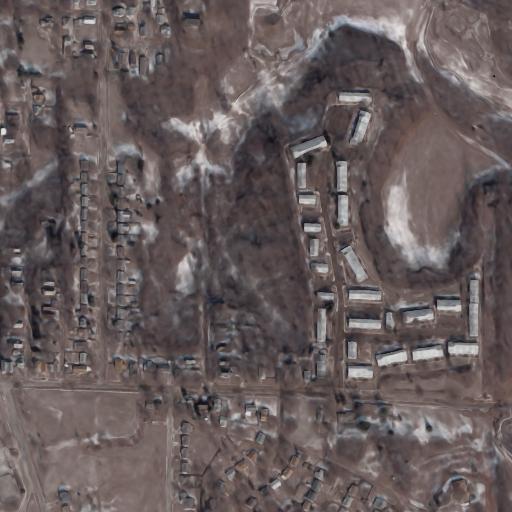}
  \end{tabularx}
\caption{Three rendered images of OMA\_374 in February with three different input sun directions, from the beginning, middle and end of February.}
\label{fig:planet-nerf-feb-changes}
\end{figure}

In Planet-NeRF, one embedding vector $\textbf{e}_m$, $m\in[1:12]$, is learnt for each month. This vector is then used to texture all RGB predictions from that month. Figure \ref{fig:planet-nerf-feb-changes} demonstrates three images rendered with the same month embedding vector, namely February, but with three different sun directions $\textbf{d}_\text{sun}$. The three directions are rendered to represent the beginning, middle, and end of February. As seen, despite all images sharing the same month embedding vector, the appearance of the rendered images still varies slightly because of the different input sun directions, showing that the Sun's influence on seasonal predictions partly remains. 

\section{Conclusion}
\label{sec:conclusion}
Addressing the dynamic nature of seasons poses a significant challenge for NeRF methodologies, encompassing complexities not only in the method and development but also in its evaluation. In this work, the seasonal predictive capability of Sat-NeRF has been explored and a simple, modular, yet effective extension has been proposed. Sat-NeRF is, acutally, already able to embed seasonal information to some extent. It does, however, get confused by the equinox months. Despite its simple nature, the proposed extension, Planet-NeRF, outperforms prior work in the case when seasonal variability is present on publicly available datasets.

In theory, the proposed extension could be integrated into any satellite NeRF 3D reconstruction architecture. At the moment, to the best of our knowledge, the only such, publicly available method is Sat-NeRF. Future work includes evaluation of the extension on other architectures. Another possible direction is to develop more accurate ground truth altitude data, possibly with different data sets for different seasons that are also matched time-wise to ground truth satellite images. Furthermore, the proposed embedding vectors can learn how an area most probable will look during a certain time of the year. They do not, however, model the unpredictability of seasonal variability, an issue left for future work.

In conclusion, the in-depth evaluation of the seasonal predictive capabilities of Sat-NeRF in combination with the proposed extension, offers promising avenues for future research in this domain.

\section{Acknowledgements}
The research was partly funded by Vinnova through the project "Framtida Människocentrerade Autonoma Regionala Fraktflygplatser", grant no. 2022-02678.

%
%
%
%

\bibliographystyle{splncs04}
\bibliography{main}

\end{document}